\documentclass[11pt]{article}

\PassOptionsToPackage{dvipsnames,table,svgnames}{xcolor}

\usepackage[preprint]{acl}

\usepackage{times}
\usepackage{latexsym}
\usepackage[T1]{fontenc}
\usepackage[utf8]{inputenc}
\usepackage{microtype}
\usepackage{inconsolata}
\usepackage{fontawesome5}

\usepackage{graphicx}
\usepackage{booktabs}
\usepackage{multirow}
\usepackage{makecell}
\usepackage{threeparttable}
\usepackage{pifont}
\usepackage{colortbl}
\usepackage{amsfonts}
\usepackage{amsmath}
\usepackage{amssymb}
\usepackage{amsthm}
\usepackage{nicefrac}
\usepackage{algorithm}
\usepackage{algpseudocode}
\usepackage{rotating}
\usepackage{siunitx}
\usepackage{adjustbox}
\usepackage{tcolorbox}
\usepackage{wrapfig}
\usepackage{enumitem}
\tcbuselibrary{skins,breakable}
\usepackage{arydshln}
\usepackage{xspace}

\renewcommand{\arraystretch}{1.08}
\definecolor{refblue}{RGB}{0, 90, 180}
\definecolor{citeblue}{RGB}{30, 110, 200}
\definecolor{urlblue}{RGB}{70, 130, 220}
\definecolor{bestcol}{RGB}{197, 214, 240}
\definecolor{secondcol}{RGB}{224, 233, 247}
\definecolor{ourbg}{RGB}{214, 234, 248}
\definecolor{ourtext}{RGB}{29, 78, 216}
\definecolor{qwenbg}{RGB}{226, 232, 240}
\definecolor{gptbg}{RGB}{226, 232, 240}
\definecolor{qwencol}{RGB}{51, 65, 85}
\definecolor{gptcol}{RGB}{51, 65, 85}
\definecolor{alfbg}{RGB}{241, 245, 249}
\definecolor{webshopbg}{RGB}{241, 245, 249}

\newcommand{\up}[1]{$_{\color{BrickRed}\uparrow\!#1}$}
\newcommand{\dn}[1]{$_{\color{MidnightBlue}\downarrow\!#1}$}
\newcommand{\second}[1]{\cellcolor{secondcol}{#1}}
\newcommand{\best}[1]{\cellcolor{bestcol}{#1}}

\newcommand{\ourmethod}{{\fontfamily{ppl}\selectfont\textsc{HyperSkill}}\xspace}

\title{{\fontfamily{ppl}\selectfont\textsc{HyperSkill}}: Self-Evolving LLM Agents via Hypergraph-Structured Skill Memory}
\author{
  \textbf{Ruiyao Xu}$^{1}$ \quad
  \textbf{Tiankai Yang}$^{2}$ \quad
  \textbf{Wei-Chieh Huang}$^{3}$ \\[0.5em]
  $^{1}$Northwestern University \\
  $^{2}$University of Southern California \\
  $^{3}$University of Illinois at Chicago
}
\begin{document}

\newcommand{\cmark}{{\color{ForestGreen}\ding{51}}}
\newcommand{\xmark}{{\color{RedOrange}\ding{55}}}

\maketitle

\begin{abstract}
As agentic tasks grow in complexity, LLM agents increasingly rely on experiential memory to reuse procedural knowledge across tasks. Effective memory design must jointly address \textit{what to store}, \textit{how is structured and retrieved}, and \textit{how memory evolves}. Existing systems tackle each only partially: they store trajectories, insights, or workflows as isolated entries, discarding compositional relationships among subtasks and reusable skills; retrieve by flat embedding similarity that ignores relational signals; and maintain memory without leveraging its relational structure. We propose \ourmethod, a hypergraph-based memory framework that jointly improves all three. \ourmethod\ represents memory as a hypergraph with two node types, subtask steps and reusable skills, where each hyperedge links the subtasks and skills from a single trajectory. Dual-path retrieval queries both subtask and trajectory levels, ranking skills by co-occurrence across retrieved trajectories. Periodic structure-informed maintenance prunes low-utility nodes and merges redundant skills via quality-weighted propagation. Across xBench, GAIA, and WebWalkerQA with \texttt{GPT-4o} and \texttt{Qwen3-30B-A3B}, \ourmethod\ outperforms ten memory baselines, yielding gains of up to $+11.51$ on GAIA and $+11.18$ on WebWalkerQA.~\footnote{Code is available at~ \href{https://github.com/rux001/HyperSkill}{\faLink}}

\end{abstract}

\section{Introduction}
\label{sec:intro}

The rapid development of LLM agents~\citep{liu-etal-2025-contextual,yao2022react,shinn2023reflexion} has significantly broadened the scope
of tasks that autonomous systems can tackle, spanning long-horizon web
navigation~\citep{chen2025xbench,gur2024a,mialon2023gaia}, interactive computer use~\citep{yang2024swe,xie2024osworld}, and multi-step scientific
reasoning~\citep{SciAgents, ren2025towards}. As tasks grow in complexity and agents operate across a continuous stream of episodes,
a single interaction is rarely sufficient to solve a problem reliably.
Agent memory systems~\citep{huang2026rethinking,hu2025memory} address this by storing and retrieving information beyond a single context window.
Among them, \textit{experiential memory}~\citep{wang2025agent, ouyang2026reasoningbank} is uniquely important: rather than encoding external facts, it distills the agent's own past trajectories into reusable knowledge, enabling \textit{self-evolving}~\citep{gao2025survey, lopez2017gradient} agents that grow more competent with each trajectory instead of starting from scratch.

\begin{figure}[t]
    \centering
    \includegraphics[width=\linewidth]{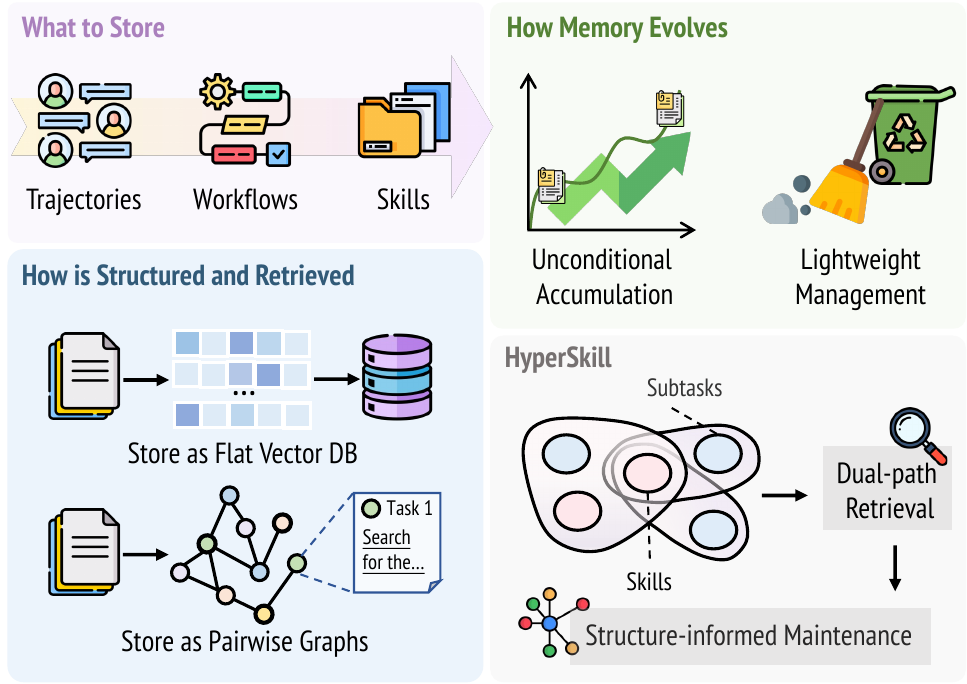}
    \caption{Three design dimensions of experiential agent memory. \ourmethod\ unifies \emph{what to store}, \emph{how memory is structured and retrieved}, and \emph{how memory evolves} via hyperedges over subtasks and skills, dual-path retrieval, and structure-informed maintenance.}
    \label{fig:intro}
    \vspace{-20pt}
\end{figure}
This has motivated a growing line of research on self-evolving agent memory~\citep{ouyang2026reasoningbank, wang2025agent, tang2025agent, zhang2025memevolve}. Early work stored raw interaction trajectories as flat records~\citep{wang2023voyager, zheng2024synapse, zhao2024expel}, while subsequent approaches shifted toward distilling experience into more compact, transferable units such as workflows~\citep{wang2025agent}, reasoning strategies~\citep{ouyang2026reasoningbank}, and execution skills~\citep{xia2026skillrl}. Despite this progress, the design of experiential memory systems remains an open problem that can be decomposed into three fundamental questions: \textit{what to store}, \textit{how is structured and retrieved}, and \textit{how memory evolves}. Existing systems address each only partially, leaving significant gaps that limit their capacity for self-improvement.

On \textit{what to store}, early systems retain raw trajectories or distill them into insights and workflows~\citep{zhao2024expel, wang2023voyager, wang2025agent}, while recent work abstracts experience into reusable skills~\citep{ouyang2026reasoningbank, zheng2025skillweaver, anthropic2024claude3}. However, these systems discard the compositional relationships among subtasks and skills. Since tasks are inherently compositional, retrieval cannot leverage shared procedural structure across superficially different tasks when these relationships are not retained. On \textit{how memory is structured and retrieved}, most approaches adopt
flat vector stores or simple binary
graphs~\citep{zhang2025g,chhikara2025mem0, rasmussen2025zep} and retrieve by embedding
similarity alone, which cannot represent nn
n-ary co-occurrence between a skill and the subtask patterns it has repeatedly succeeded with.
On \textit{how memory evolves}, the majority of systems accumulate knowledge
without any curation~\citep{wang2023voyager, zhao2024expel}, and the few
that do introduce only lightweight operations such as pruning or
deduplication~\citep{fang2025memp,zheng2025skillweaver}, applied
independently to each node.

These gaps share a common \emph{structural root}: a trajectory is not a
single unit of knowledge but an $n$-ary association among subtasks, skills,
and an outcome, structure that flat lists and pairwise edges represent
only lossily. Hypergraphs~\citep{zhou2006learning,tang2025trainingfree}
preserve such associations natively by letting each hyperedge connect
more than two nodes simultaneously. We propose \ourmethod, which organizes experiential memory as a hypergraph
$\mathcal{G}=(\mathcal{V},\mathcal{E})$ with subtask nodes
$\mathcal{V}_u$ and skill nodes $\mathcal{V}_s$, where each hyperedge
groups the subtasks and skills from a single trajectory together with a
distilled lesson and a utility score. This design directly improves the
three gaps: \ding{182}~hyperedges preserve the full
compositional context that flat stores discard; \ding{183}~skill nodes are
shared across hyperedges, enabling us to rank a candidate skill by how
frequently it appears across the retrieved trajectories, a structural
signal that semantic similarity alone cannot provide; and
\ding{184}~quality-weighted propagation over the hypergraph enables merging
redundant skills by jointly considering node utility and neighborhood
topology. Across three agentic benchmarks (xBench, GAIA, and WebWalkerQA) under two different model families, \ourmethod\ consistently
outperforms ten memory baselines.

\section{Related Work}
\label{sec:related_works}

\textbf{Memory Mechanisms in Agent Systems.}
LLM agents are increasingly equipped with external memory to overcome fixed context windows~\citep{huang2026rethinking}, serving roles such as factual memory~\citep{gutierrez2024hipporag, rasmussen2025zep}, working memory~\citep{packer2023memgpt}, and experiential memory, which distills the agent's own trajectories into reusable knowledge for self-improvement. We focus on the latter and categorize prior work along three dimensions, identifying a gap in each.
\textit{(i)~What is stored:}
Early works retain raw trajectories~\citep{zheng2024synapse, wen2023dilu}, which preserve context but carry noise; later approaches abstract trajectories into insights or workflows~\citep{zhao2024expel, ouyang2026reasoningbank, wang2025agent, fang2025memp}, gaining generality at the cost of procedural detail; recent work distills reusable skills~\citep{zheng2025skillweaver, wang2023voyager, xia2026skillrl} but treats each skill in isolation. None preserves reusable subtask decompositions, limiting transfer of compositional planning across episodes.
\textit{(ii)~How memory is structured and retrieved:}
Most systems use flat vector or JSON stores with embedding-based retrieval~\citep{zhao2024expel, ouyang2026reasoningbank, wang2025agent, zheng2025skillweaver}; a few adopt graph structures~\citep{zhang2025g, yang2026plugmem} to capture relational dependencies, but pairwise edges decompose nn
n-ary co-occurrence into disconnected binary links and discard the joint compositional context of an episode.
\textit{(iii)~How memory evolves:}
Most systems accumulate unconditionally~\citep{wang2023voyager, zhao2024expel, wang2025agent}; a few add lightweight per-entry pruning~\citep{zheng2025skillweaver}, consolidation~\citep{zhang2025g}, or deduplication~\citep{tang2025agent}, but none leverages structural signals or jointly considers quality and graph topology. A more detailed discussion is in Appendix~\ref{app:related_works}.

and retrieved, and \emph{how} memory evolves over time, identifying a gap in
each.
\textit{(i)~What is stored:}
Early works store raw trajectories of entire
episodes~\citep{zheng2024synapse, wen2023dilu}, which preserve full context
but include substantial noise; later approaches abstract experience into
insights and workflows~\citep{zhao2024expel, ouyang2026reasoningbank,
wang2025agent, fang2025memp}, improving generalizability but discarding
fine-grained procedural detail; while others distil reusable skills from
trajectories~\citep{zheng2025skillweaver, wang2023voyager, xia2026skillrl},
retaining concise knowledge units but treating each skill in isolation
without compositional structure. Existing systems don't maintain
reusable subtask decompositions that capture what procedural sequence was
used to solve a task, limiting the agent's ability to transfer compositional
planning knowledge across episodes.
\textit{(ii)~How memory is structured and retrieved:}
Most systems store memory as flat vector databases or JSON files and rely
solely on semantic similarity for
retrieval~\citep{zhao2024expel, ouyang2026reasoningbank, wang2025agent,
zheng2025skillweaver}, limiting both the relevance and comprehensiveness of
retrieved knowledge. A few works explore graph structures to capture
relational dependencies~\citep{zhang2025g, yang2026plugmem, xu2026gnnasjudge}, enabling
structural traversal, but rely on pairwise edges that decompose $n$-ary
co-occurrence into disconnected binary links and discard the joint
compositional context of an episode.
\textit{(iii)~How memory evolves:}
Most systems accumulate memory
unconditionally~\citep{wang2023voyager, zhao2024expel, wang2025agent};
a few add lightweight management such as
pruning~\citep{zheng2025skillweaver},
consolidation~\citep{zhang2025g}, or
deduplication~\citep{tang2025agent}, but all operate on individual entries
without leveraging structural information, with no maintenance guided
jointly by quality signals and graph topology. A more detailed discussion is provided in Appendix~\ref{app:related_works}.

\section{\ourmethod}
\label{sec:method}
\begin{figure*}[t]
    \centering
    \includegraphics[width=\textwidth]{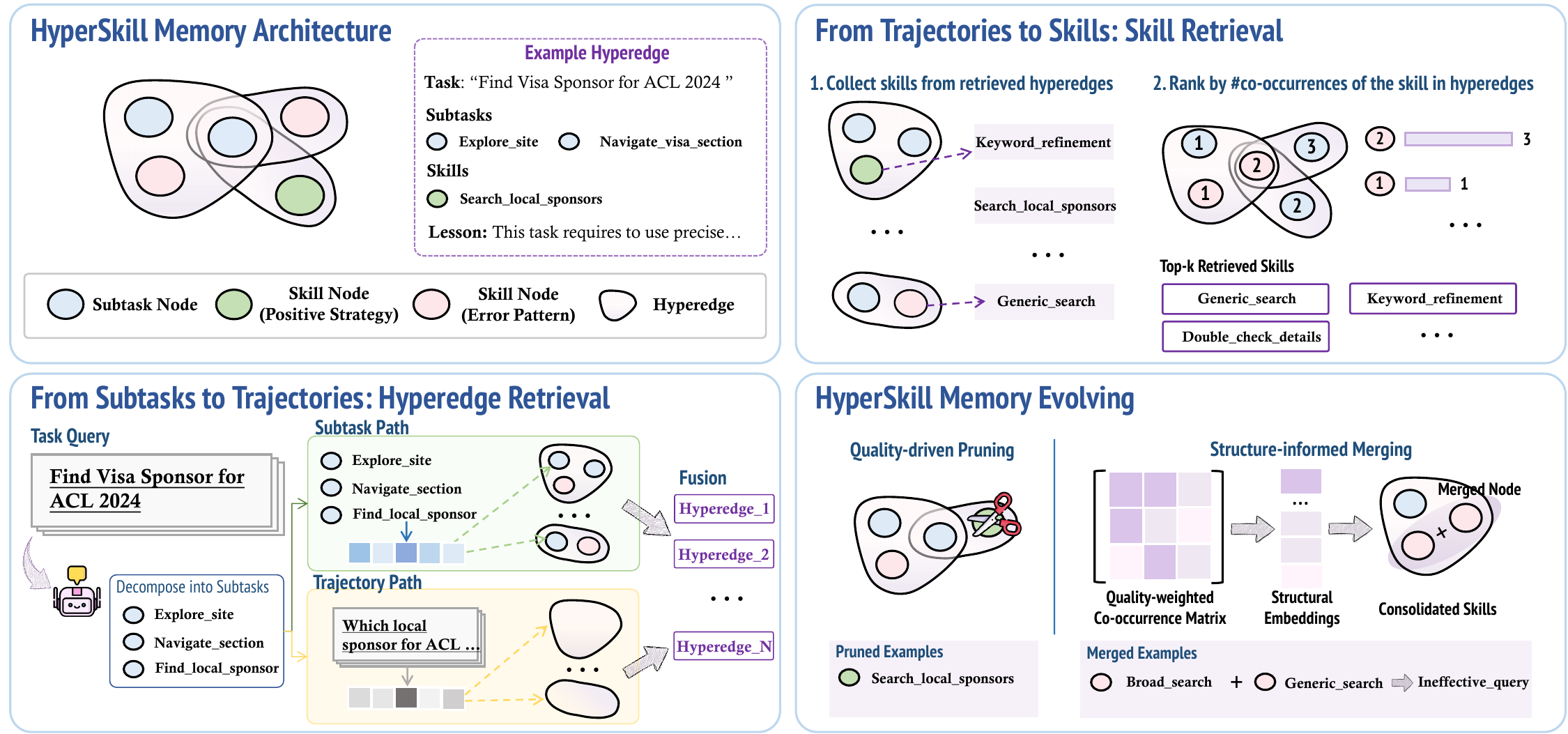}
    \caption{Overview of \ourmethod. Given a new task, the agent
    decomposes it into subtasks, retrieves relevant trajectories via
    dual-path hyperedge retrieval, ranks skills by co-occurrence, and
    executes with the retrieved context. After task completion, new
    nodes and a hyperedge are added to the memory, which periodically
    undergoes pruning and merging.}
    \vspace{-10pt}
    \label{fig:architecture}
\end{figure*}

\subsection{Preliminaries}
\label{sec:prelim}

\textbf{Agent, Environment, and Memory.}
Following previous work~\citep{zhang2025memevolve}, we formalize an LLM-based agent system as a policy $\pi_\theta$ interacting with an environment over
trajectories $\tau_i = (d_i, o_i, a_i, r_i)$, where $d_i$ is a task description, $o_i$ and $a_i$
are observation and action sequences, and $r_i \in \{\texttt{success}, \texttt{fail}\}$ is the
outcome. At each step $t$, the agent selects
$a_t \sim \pi_\theta(\cdot \mid o_t, m_t)$,
where $m_t$ is a memory context retrieved from an external memory module $\mathcal{M}$.
Given growing history $\mathcal{H} = \{\tau_1, \dots, \tau_N\}$, $\mathcal{M}$ supports
retrieval $m_t = g_{\mathcal{M}}(d, \mathcal{H})$ and update $\mathcal{M} \leftarrow \textsc{Update}(\mathcal{M}, \tau_i)$,
with the objective of maximizing expected task performance:
\begin{equation}
    \max_{\mathcal{M}} \;
    \mathbb{E}_{d \sim \mathcal{D}}
    \left[
    R\!\left(
    \pi_\theta(\cdot \mid d, g_{\mathcal{M}}(d, \mathcal{H}))
    \right)
    \right].
\end{equation}

\subsection{\ourmethod\ Memory Architecture}
\label{sec:architecture}

A hypergraph~\citep{zhou2006learning} $\mathcal{G} = (\mathcal{V}, \mathcal{E})$
generalizes ordinary graphs by allowing each hyperedge
$e \in \mathcal{E} \subseteq 2^{\mathcal{V}}$ to connect an arbitrary subset of nodes
($|e| \geq 2$), capturing higher-order $n$-ary relationships among multiple nodes that pairwise
edges cannot represent.
\ourmethod\ exploits this structure to encode both compositional task decompositions
and reusable execution knowledge in a single unified memory.

\noindent[\ding{226}]~\textbf{Nodes.}
Each $v_i \in \mathcal{V}$ is a discrete, reusable unit of task knowledge distilled
from agent experience. The node set partitions as $\mathcal{V} = \mathcal{V}_u \cup \mathcal{V}_s$,
where each node can be represented by a tuple:
\begin{equation}
    v_i \;\triangleq\; \bigl(c_i,\;\ell_i,\;\gamma_i\bigr), \qquad \ell_i \in \{\texttt{u},\,\texttt{s}\},
    \label{eq:node}
\end{equation}
with $c_i$ as natural-language content, $\ell_i$ as type label,
and $\gamma_i \in [0,1]$ as \emph{utility score}, a dynamic measure that tracks empirical success rate and step efficiency.
Concretely, for all nodes and hyperedges that participated in retrieval during
trajectory $\tau_i$, we increment $\sigma(\cdot)$ on success and update $\bar{T}(\cdot)$
with step count $T_i$, then recompute:
\begin{equation}
    \gamma(\cdot) = \beta \cdot \frac{\sigma(\cdot)}{\nu(\cdot)}
    + (1 - \beta) \cdot \left(1 - \frac{\bar{T}(\cdot) - T_{\min}}
    {T_{\max} - T_{\min}}\right),
    \label{eq:gamma}
\end{equation}
where $\nu(\cdot)$ is the total retrieval count, $\sigma(\cdot)$ the success count,
$\bar{T}(\cdot)$ the running average step count, and $\beta$ balances success rate
against step efficiency. The two node types reflect complementary
sides of task knowledge: \emph{subtask nodes} encode the compositional
workflow structure of a task, while \emph{skill nodes} encode reusable
execution strategies that transfer across tasks.
\begin{itemize}[label=$\circ$, leftmargin=*, itemsep=0.1pt, topsep=0.2pt]
    \item \textbf{Subtask nodes} $\mathcal{V}_u$: each $v_u \triangleq (c_u, \texttt{u}, \gamma_u)$
    captures one step in a task decomposition.
    The content $c_u$ encodes a step description together with any
    preconditions under which it applies.
    \item \textbf{Skill nodes} $\mathcal{V}_s$: each $v_s \triangleq (c_s, \texttt{s}, \gamma_s)$
    captures reusable knowledge extracted via
    outcome-conditioned prompting.
    The content $c_s$ encodes a concise strategy label and a description covering what to do, when to apply it, and why it works (or fails).
    Successful trajectories yield \emph{positive strategies} to follow;
    failed trajectories yield \emph{error patterns} to avoid.
\end{itemize}

\noindent[\ding{226}]~\textbf{Hyperedges.}
Each hyperedge $e_j \in \mathcal{E}$ corresponds to a single trajectory $\tau_j$
and is defined as a tuple
$e_j \triangleq (V_j,\, d_j,\, \ell_j,\, \gamma_j)$,
where $V_j = V_j^u \cup V_j^s$, with
$V_j^u \subseteq \mathcal{V}_u$ the subtask nodes and
$V_j^s \subseteq \mathcal{V}_s$ the skill nodes extracted from $\tau_j$;
$d_j$ is the original task description;
and $\ell_j$ is a concise \emph{distilled lesson} capturing the transferable
pattern the trajectory reveals.
For retrieval, each hyperedge is represented by the embedding
$\mathbf{h}_e = \boldsymbol{\phi}(d_j \,\|\, \ell_j)$, which concatenates the
task description with the distilled lesson so that similarity matching
captures both surface wording and the underlying transferable pattern.

\subsection{From Subtasks to Trajectories: Hyperedge Retrieval}
\label{sec:episodic_retrieval}

Prior methods explore diverse forms of experiential knowledge, from raw
trajectories~\citep{zheng2024synapse, wen2023dilu} and distilled
insights~\citep{ouyang2026reasoningbank, zhao2024expel} to reusable
skills~\citep{zheng2025skillweaver, wang2023voyager}, but most rely on the
full task description as the sole retrieval query. This misses trajectories that
are superficially different yet share similar procedural subtask patterns.
To retrieve more comprehensive trajectory-level context, \ourmethod\ queries the
hypergraph through two complementary paths: one decomposes the current task
into subtasks and matches them against stored subtask nodes, surfacing
trajectories with shared procedural structure regardless of task-level wording;
the other matches the task description directly against hyperedge
representations that encode both the original task and its distilled lesson.

\noindent\textbf{Task decomposition and subtask path.}
Upon receiving a new task $d_q$, the agent first invokes an LLM call to
produce a fine-grained decomposition
$\mathcal{P}_0 = \{d_1, \ldots, d_m\}$, where each $d_i$ describes a subtask unit.
This decomposition mirrors the subtask structure already stored in the
hypergraph, enabling retrieval at the subtask level rather than the
full-task level.
We encode the full decomposition as a single vector
$\mathbf{h}_{\mathcal{P}_0} = \boldsymbol{\phi}(\mathcal{P}_0)$ using a
fixed-length text encoder $\boldsymbol{\phi}$, and retrieve the top-$k_u$
subtask nodes whose embeddings
$\mathbf{h}_v = \boldsymbol{\phi}(c_v)$ are most similar:
\begin{equation}
    R_u = \underset{v \in \mathcal{V}_u}{\operatorname{top\text{-}}k_u}
    \;\operatorname{sim}\!\bigl(\mathbf{h}_{\mathcal{P}_0},\,\mathbf{h}_v\bigr).
    \label{eq:subtask_retrieval}
\end{equation}
Because each subtask node is incident to one or more hyperedges, the
matched nodes naturally expand to a set of candidate trajectories:
$\mathcal{E}_{\mathrm{sub}} = \{e \in \mathcal{E} \mid
V_e \cap R_u \neq \varnothing\}$,
where $V_e$ denotes the node set of hyperedge $e$.
This path is particularly effective at surfacing trajectories that share
procedural substructure with the current task, even when their
task-level descriptions differ substantially.

\noindent\textbf{Trajectory path and fusion.}
The subtask path alone may miss trajectories whose overall task descriptions
are highly relevant but whose individual subtask nodes were not among the
top-$k_u$ matches.
A complementary trajectory-level path addresses this by encoding the
full task description as $\mathbf{h}_{d_q} = \boldsymbol{\phi}(d_q)$ and
ranking hyperedges directly by similarity to the query:
\begin{equation}
    \mathcal{E}_{\mathrm{traj}} =
    \underset{e \in \mathcal{E}}{\operatorname{top\text{-}}k_e}
    \;\operatorname{sim}\!\bigl(\mathbf{h}_{d_q},\,\mathbf{h}_{e}\bigr),
    \label{eq:task_retrieval}
\end{equation}
where $\mathbf{h}_e = \boldsymbol{\phi}(d_e \,\|\, \ell_e)$ concatenates the
original task description with the distilled lesson, so that ranking
captures both surface wording and the underlying transferable pattern.
Candidates from both paths are merged into a unified trajectory set:
\begin{equation}
    \mathcal{E}^{*} = \mathcal{E}_{\mathrm{sub}} \cup
    \mathcal{E}_{\mathrm{traj}}.
    \label{eq:fusion}
\end{equation}
Together, the two paths are complementary: the subtask path retrieves
trajectories with shared procedural substructure, while the trajectory
path retrieves those with similar high-level task semantics.
The distilled lessons $\mathcal{L}_q = \{\ell_e \mid e \in \mathcal{E}^{*}\}$
from the fused set are provided to the agent as trajectory-level context
during execution.

\subsection{From Trajectories to Skills: Skill Retrieval}
\label{sec:skill_retrieval}

Given the fused trajectory set $\mathcal{E}^{*}$, the next step is to select
which skills to surface as execution guidance. Each hyperedge in
$\mathcal{E}^{*}$ contains the skill nodes extracted from that trajectory,
so the candidate pool is their union:
\begin{equation}
    R_s = \bigcup_{e \in \mathcal{E}^{*}} V_e^s \;\subseteq\; \mathcal{V}_s,
    \label{eq:skill_candidates}
\end{equation}
where $V_e^s$ denotes the skill nodes of hyperedge $e$.
A flat memory system would rank these candidates by embedding similarity to
$d_q$ alone. However, semantic similarity only measures whether a skill's
description \emph{looks like} the current task; it does not capture whether
the skill has actually been applied in trajectories relevant to the current
task.

\noindent\textbf{Co-occurrence ranking.}
\ourmethod\ exploits the fact that skill nodes, unlike subtask nodes,
are deduplicated and reused across trajectories. A skill that appears in
many of the retrieved trajectories has been repeatedly applied in
task-relevant contexts and is therefore more likely to be useful. We
formalize this with a simple co-occurrence count: for each candidate
$v \in R_s$, we count how many retrieved hyperedges contain $v$:
\begin{equation}
    \kappa(v) = \bigl|\bigl\{\, e \in \mathcal{E}^* \;\big|\;
    v \in V_e \,\bigr\}\bigr|.
    \label{eq:cooc}
\end{equation}
The top-$k_s$ skills by $\kappa$ are selected as execution guidance,
with ties broken by embedding similarity to $d_q$:
\begin{equation}
    \mathcal{S}_q = \underset{v \in R_s}{\operatorname{top\text{-}}k_s}
    \; \kappa(v).
    \label{eq:skill_retrieve}
\end{equation}

\noindent\textbf{Execution and memory update.}
The retrieved context $(\mathcal{S}_q, \mathcal{L}_q)$ is assembled once per task and held
fixed throughout execution, instantiating the memory context
$m_t = (\mathcal{S}_q, \mathcal{L}_q)$. At each step $t$, the agent
selects an action conditioned on the current observation $o_t$, the
retrieved skills $\mathcal{S}_q$, the distilled lessons $\mathcal{L}_q$, and a sliding window
of recent interaction history:
\begin{equation}
    a_t \sim \pi_\theta\!\bigl(
    \cdot \;\big|\;
    o_t,\; \mathcal{S}_q,\; \mathcal{L}_q,\; \mathrm{hist}_{t-w:t-1}
    \bigr).
    \label{eq:act}
\end{equation}
Upon task completion with outcome $r_i$ and step count $T_i$, an
outcome-conditioned LLM call extracts skill nodes $S_i^{\mathrm{new}}$ and a
distilled lesson $\ell_{e_i}$ from the full trajectory. Subtask nodes
$U_i^{\mathrm{new}}$ are taken directly from $\mathcal{P}_0$. Before inserting each
newly extracted skill node into $\mathcal{V}_s$, we compute its embedding
similarity to all existing skill nodes; if the similarity exceeds a
threshold $\delta_{\mathrm{dedup}}$, the new node is merged into its nearest existing
neighbor rather than added as a duplicate, keeping the skill vocabulary
compact and reusable. A new hyperedge over $U_i^{\mathrm{new}} \cup S_i^{\mathrm{new}}$ is then
added to $\mathcal{E}$ with lesson $\ell_{e_i}$ and task description $d_q$;
its embedding is $\boldsymbol{\phi}(d_q \,\|\, \ell_{e_i})$, matching
Eq.~\ref{eq:task_retrieval}. The utility score $\gamma(\cdot)$ is
recomputed via Eq.~\ref{eq:gamma} for all elements that participated in
retrieval.

\subsection{\ourmethod\ Memory Evolving}
\label{sec:maintenance}

As \ourmethod\ accumulates tasks, $\mathcal{G}$ risks growing stale: nodes may
become semantically redundant, and repeatedly retrieved but low-quality knowledge can
actively mislead future decisions.
To keep memory compact, non-redundant, and high-quality, \ourmethod\ periodically
refines $\mathcal{G}$ every $N_{\mathrm{maint}}$ tasks via two operations: \emph{quality-driven pruning} and \emph{structure-informed merging}.

\noindent\textbf{Quality-driven pruning.}
Nodes that have been retrieved sufficiently often but consistently fail to improve
agent outcomes are explicitly removed from $\mathcal{G}$:
\begin{equation}
\begin{aligned}
    \mathcal{V} \;\leftarrow\; \mathcal{V} \setminus
    \bigl\{ v \in \mathcal{V} \,\big|\, \nu(v) \geq N_{\min} \\
    \wedge\; \gamma(v) < \tau_{\mathrm{prune}} \bigr\},
\end{aligned}
\label{eq:prune}
\end{equation}
where $\nu(v)$ is the total number of times $v$ has been retrieved and
$\gamma(v)$ is its composite quality score (Eq.~\ref{eq:gamma}), reflecting
empirical success rate and step efficiency accumulated over those retrievals. When a node is deleted, it is removed from all hyperedge memberships.

\begin{table*}[t]
\centering
\small
\caption{
  Main results across three benchmarks.
  We report success rate (SR\,\%), average number of steps (\#Steps), and average tool calls (\#Calls) per task.
  \colorbox{bestcol}{\textbf{Best}};
  \colorbox{secondcol}{2nd best} per backbone.
  {$_{\color{RedOrange}\uparrow}$}/{$_{\color{BlueGreen}\downarrow}$}: change vs.\ No Memory.
  For \#Steps and \#Calls, lower is better.
}
\label{tab:main_results}
\resizebox{\linewidth}{!}{%
\renewcommand{\arraystretch}{1.15}
\begin{tabular}{cl|c:cc|c:cc|c:cc}
\toprule
\textbf{Backbone} & \textbf{Method}
  & \multicolumn{3}{c|}{\textbf{xBench}}
  & \multicolumn{3}{c|}{\textbf{GAIA}}
  & \multicolumn{3}{c}{\textbf{WebWalkerQA}} \\
\cmidrule(lr){3-5}\cmidrule(lr){6-8}\cmidrule(lr){9-11}
& & SR & \#Steps & \#Calls & SR & \#Steps & \#Calls & SR & \#Steps & \#Calls \\
\midrule

\multirow{12}{*}{%
  \rotatebox{90}{%
    \makecell{%
      \includegraphics[height=14pt]{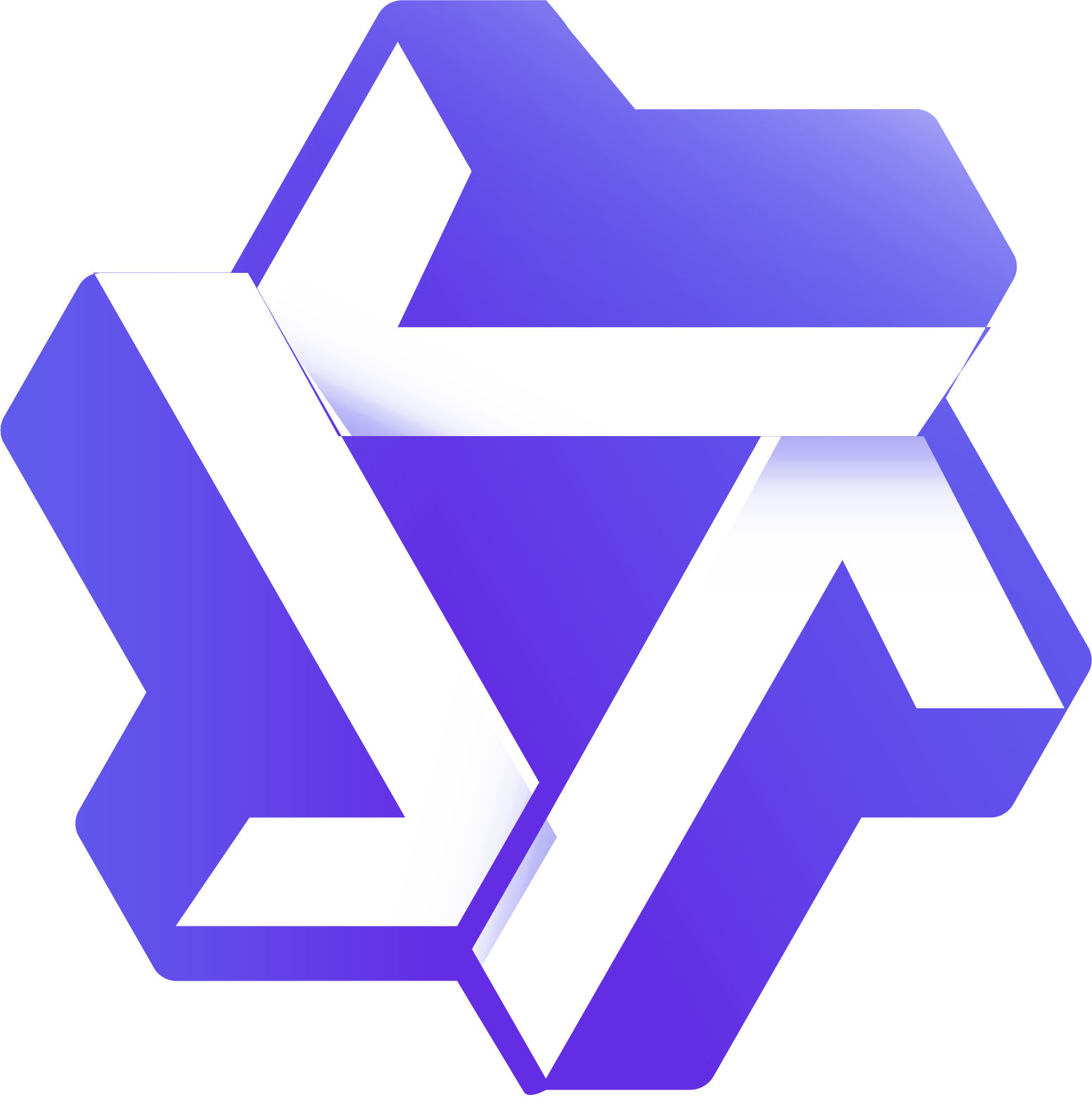}\\[1pt]
      \colorbox{qwenbg}{%
        \hspace{3pt}\texttt{\textcolor{black}{\textbf{Qwen3-30B-A3B}}}\hspace{3pt}%
      }%
    }%
  }%
}
& No Memory
  & 46.00\up{0.00} & 4.87\up{0.00} & 5.05\up{0.00}
  & 32.12\up{0.00} & 8.23\up{0.00} & 4.75\up{0.00}
  & 39.41\up{0.00} & 4.07\up{0.00} & 3.87\up{0.00} \\
\cmidrule{2-11}
& ReasoningBank
  & 43.00\dn{3.00} & 4.47\dn{0.40} & \second{4.72\dn{0.33}}
  & 29.70\dn{2.42} & 7.44\dn{0.79} & 4.42\dn{0.33}
  & 44.71\up{5.30} & 3.64\dn{0.43} & 3.72\dn{0.15} \\
& ExpeL
  & 45.00\dn{1.00} & 5.14\up{0.27} & 6.06\up{1.01}
  & 33.33\up{1.21} & 7.38\dn{0.85} & 4.15\dn{0.60}
  & \second{47.06\up{7.65}} & 4.04\dn{0.03} & 3.94\up{0.07} \\
& AWM
  & 44.00\dn{2.00} & 4.86\dn{0.01} & 5.49\up{0.44}
  & 32.73\up{0.61} & 7.07\dn{1.16} & \second{3.92\dn{0.83}}
  & 41.76\up{2.35} & 3.84\dn{0.23} & 3.72\dn{0.15} \\
& Generative
  & 49.00\up{3.00} & 4.46\dn{0.41} & 4.97\dn{0.08}
  & \second{34.55\up{2.43}} & 7.43\dn{0.80} & 3.97\dn{0.78}
  & 41.18\up{1.77} & 3.82\dn{0.25} & 3.87\up{0.00} \\
& Voyager
  & 50.00\up{4.00} & 4.77\dn{0.10} & 5.05\up{0.00}
  & 34.55\up{2.43} & 7.47\dn{0.76} & 4.16\dn{0.59}
  & 46.47\up{7.06} & 3.84\dn{0.23} & 3.82\dn{0.05} \\
& DILU
  & \second{52.00\up{6.00}} & 5.01\up{0.14} & 5.45\up{0.40}
  & 32.12\up{0.00} & 7.14\dn{1.09} & 4.17\dn{0.58}
  & 44.12\up{4.71} & 3.87\dn{0.20} & 3.95\up{0.08} \\
& Cheatsheet
  & 37.00\dn{9.00} & \best{4.03\dn{0.84}} & 4.84\dn{0.21}
  & 27.88\dn{4.24} & \second{6.84\dn{1.39}} & 5.28\up{0.53}
  & 39.41\up{0.00} & \second{3.62\dn{0.45}} & 4.06\up{0.19} \\
& MemP
  & 45.00\dn{1.00} & \second{4.42\dn{0.45}} & \best{4.24\dn{0.81}}
  & 32.73\up{0.61} & 6.97\dn{1.26} & 4.42\dn{0.33}
  & 45.88\up{6.47} & 3.72\dn{0.35} & \second{3.66\dn{0.21}} \\
& MemEvolve
  & 39.00\dn{7.00} & 4.64\dn{0.23} & 4.72\dn{0.33}
  & 28.48\dn{3.64} & \best{5.08\dn{3.15}} & \best{3.67\dn{1.08}}
  & 37.06\dn{2.35} & \best{3.61\dn{0.46}} & \best{3.32\dn{0.55}} \\
& PlugMem
  & 42.00\dn{4.00} & 5.07\up{0.20} & 5.18\up{0.13}
  & 34.55\up{2.43} & 7.02\dn{1.21} & 4.18\dn{0.57}
  & 35.88\dn{3.53} & 3.96\dn{0.11} & 3.83\dn{0.04} \\
\cmidrule{2-11}
& \ourmethod
  & \best{52.00\up{6.00}} & 4.52\dn{0.35} & 4.88\dn{0.17}
  & \best{36.97\up{4.85}} & 7.25\dn{0.98} & 5.48\up{0.73}
  & \best{50.59\up{11.18}} & 4.04\dn{0.03} & 4.06\up{0.19} \\

\midrule

\multirow{12}{*}{%
  \rotatebox{90}{%
    \makecell{%
      \includegraphics[height=14pt]{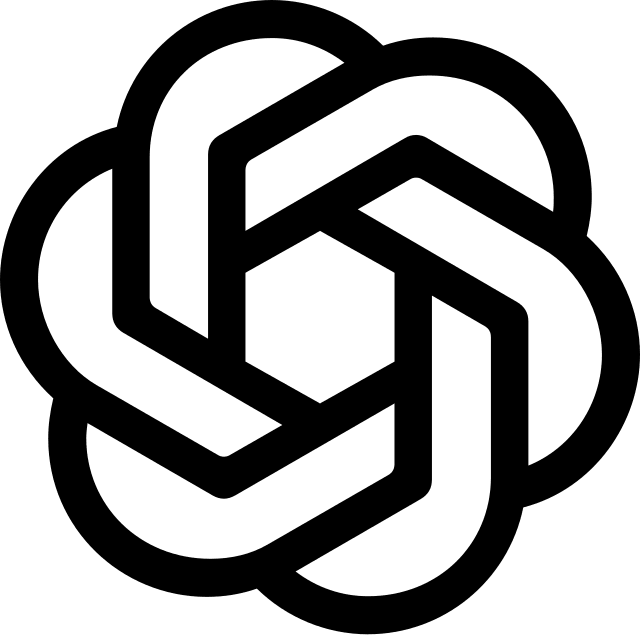}\\[1pt]
      \colorbox{gptbg}{%
        \hspace{3pt}\texttt{\textcolor{black}{\textbf{GPT-4o}}}\hspace{3pt}%
      }%
    }%
  }%
}
& No Memory
  & 54.00\up{0.00} & 6.29\up{0.00} & 6.33\up{0.00}
  & 32.73\up{0.00} & 6.59\up{0.00} & 8.23\up{0.00}
  & 46.47\up{0.00} & 5.82\up{0.00} & 7.04\up{0.00} \\
\cmidrule{2-11}
& ReasoningBank
  & 57.00\up{3.00} & 8.22\up{1.93} & 12.27\up{5.94}
  & 33.33\up{0.60} & 7.21\up{0.62} & 10.17\up{1.94}
  & 47.65\up{1.18} & \second{5.59\dn{0.23}} & 6.45\dn{0.59} \\
& ExpeL
  & 54.00\up{0.00} & \best{5.27\dn{1.02}} & \best{6.11\dn{0.22}}
  & 35.76\up{3.03} & 6.63\up{0.04} & 9.48\up{1.25}
  & 47.06\up{0.59} & \best{5.58\dn{0.24}} & \second{6.32\dn{0.72}} \\
& AWM
  & 52.00\dn{2.00} & 6.47\up{0.18} & 8.12\up{1.79}
  & 35.15\up{2.42} & 7.13\up{0.54} & 9.08\up{0.85}
  & 47.65\up{1.18} & 5.77\dn{0.05} & 6.49\dn{0.55} \\
& Generative
  & 56.00\up{2.00} & 7.65\up{1.36} & 11.34\up{5.01}
  & 36.36\up{3.63} & 6.39\dn{0.20} & 7.68\dn{0.55}
  & \best{51.18\up{4.71}} & 5.77\dn{0.05} & 6.33\dn{0.71} \\
& Voyager
  & 47.00\dn{7.00} & 7.99\up{1.70} & 11.27\up{4.94}
  & 31.52\dn{1.21} & 6.52\dn{0.07} & 9.15\up{0.92}
  & 48.82\up{2.35} & 6.07\up{0.25} & 6.93\dn{0.11} \\
& DILU
  & 53.00\dn{1.00} & 7.16\up{0.87} & 9.98\up{3.65}
  & 33.33\up{0.60} & \best{5.34\dn{1.25}} & \second{5.63\dn{2.60}}
  & 48.24\up{1.77} & 6.05\up{0.23} & 6.97\dn{0.07} \\
& Cheatsheet
  & 55.00\up{1.00} & 7.56\up{1.27} & 10.92\up{4.59}
  & 35.15\up{2.42} & 6.27\dn{0.32} & 8.36\up{0.13}
  & 47.06\up{0.59} & 6.02\up{0.20} & 7.25\up{0.21} \\
& MemP
  & 48.00\dn{6.00} & 7.59\up{1.30} & 10.76\up{4.43}
  & 33.33\up{0.60} & 7.04\up{0.45} & 8.64\up{0.41}
  & \second{50.59\up{4.12}} & 5.66\dn{0.16} & \best{6.03\dn{1.01}} \\
& MemEvolve
  & 47.00\dn{7.00} & 6.68\up{0.39} & 9.06\up{2.73}
  & 36.36\up{3.63} & 6.97\up{0.38} & 7.29\dn{0.94}
  & 48.24\up{1.77} & 6.09\up{0.27} & 7.34\up{0.30} \\
& PlugMem
  & \second{59.00\up{5.00}} & 5.94\dn{0.35} & \second{6.22\dn{0.11}}
  & \second{41.94\up{9.21}} & \second{5.73\dn{0.86}} & \best{5.11\dn{3.12}}
  & 47.06\up{0.59} & 5.96\up{0.14} & 6.49\dn{0.55} \\
\cmidrule{2-11}
& \ourmethod
  & \best{62.00\up{8.00}} & \second{5.27\dn{1.02}} & 6.52\up{0.19}
  & \best{44.24\up{11.51}} & 6.01\dn{0.58} & 7.35\dn{0.88}
  & \best{51.18\up{4.71}} & 5.93\up{0.11} & 7.04\up{0.00} \\

\bottomrule
\end{tabular}
}%
\end{table*}

\noindent\textbf{Structure-informed merging.} Skill nodes that exhibit strong quality-weighted co-occurrence \emph{and} encode
semantically similar execution knowledge are candidates for consolidation into a
single higher-order skill.
Subtask nodes $\mathcal{V}_u$ are excluded, as merging planning steps would
corrupt the compositional structure of task decompositions.
To identify merge candidates within $\mathcal{V}_s$, we compute
\emph{quality-weighted structural embeddings}.
We first construct a co-occurrence matrix
$\mathbf{W} \in \mathbb{R}^{|\mathcal{V}_s| \times |\mathcal{V}_s|}$:
\begin{equation}
    W_{ij} = \min\!\bigl(\gamma(v_i),\,\gamma(v_j)\bigr)
             \cdot \sum_{\substack{e \in \mathcal{E} \\ v_i \in V_e,\, v_j \in V_e}} \frac{1}{|V_e|},
    \label{eq:W}
\end{equation}
where $1/|V_e|$ normalises by hyperedge size following the homophily
assumption~\citep{tang2025trainingfree}---nodes sharing a small, tight hyperedge
are more likely to encode similar knowledge than those sharing a large one---and
$\min(\gamma(v_i), \gamma(v_j))$ ensures that high edge weights occur only when \emph{both} nodes are useful.
Structural embeddings are then obtained via $L$-step symmetric propagation
following~\citep{tang2025trainingfree}:
\begin{equation}
\begin{aligned}
    \tilde{\mathbf{Z}} &= \mathbf{S}\,\mathbf{Z}, \\
    \mathbf{S} &= (1-\alpha)^{L}\,\hat{\mathbf{W}}^{L}
               + \alpha \sum_{l=0}^{L-1}(1-\alpha)^{l}\,\hat{\mathbf{W}}^{l}, \\
    \hat{\mathbf{W}} &= \hat{\mathbf{D}}^{-\nicefrac{1}{2}}
                       \bigl(\mathbf{W}+\mathbf{I}\bigr)\,
                       \hat{\mathbf{D}}^{-\nicefrac{1}{2}},
\end{aligned}
\label{eq:propagation}
\end{equation}
where $\mathbf{Z} \in \mathbb{R}^{|\mathcal{V}_s| \times d}$ stacks plain skill node embeddings,
$\hat{\mathbf{D}}$ is the degree matrix of $(\mathbf{W}+\mathbf{I})$,
$\alpha \in (0,1)$ balances individual node identity against neighbourhood context,
and $L$ controls the propagation depth.
Since node utility is already encoded in $\mathbf{W}$, the propagated embeddings
$\tilde{\mathbf{Z}}$ jointly reflect structural co-occurrence and skill quality,
and a pair $(v_a, v_b)$ is selected as a merge candidate by similarity alone:
\begin{equation}
    \mathrm{sim}\!\left(\tilde{\mathbf{z}}_a,\,\tilde{\mathbf{z}}_b\right) \geq \delta_{\mathrm{merge}}.
    \label{eq:merge_cand}
\end{equation}
For each selected pair, the agent reasons over the contents of $v_a$, $v_b$,
and their shared trajectory fragments to propose a merged node $v_{\mathrm{new}}$
that consolidates their knowledge into a higher-order skill.
All hyperedge memberships are reassigned to $v_{\mathrm{new}}$,
preserving structural integrity across $\mathcal{G}$.

\begin{figure*}[t]
    \centering
    \includegraphics[width=0.95\linewidth]{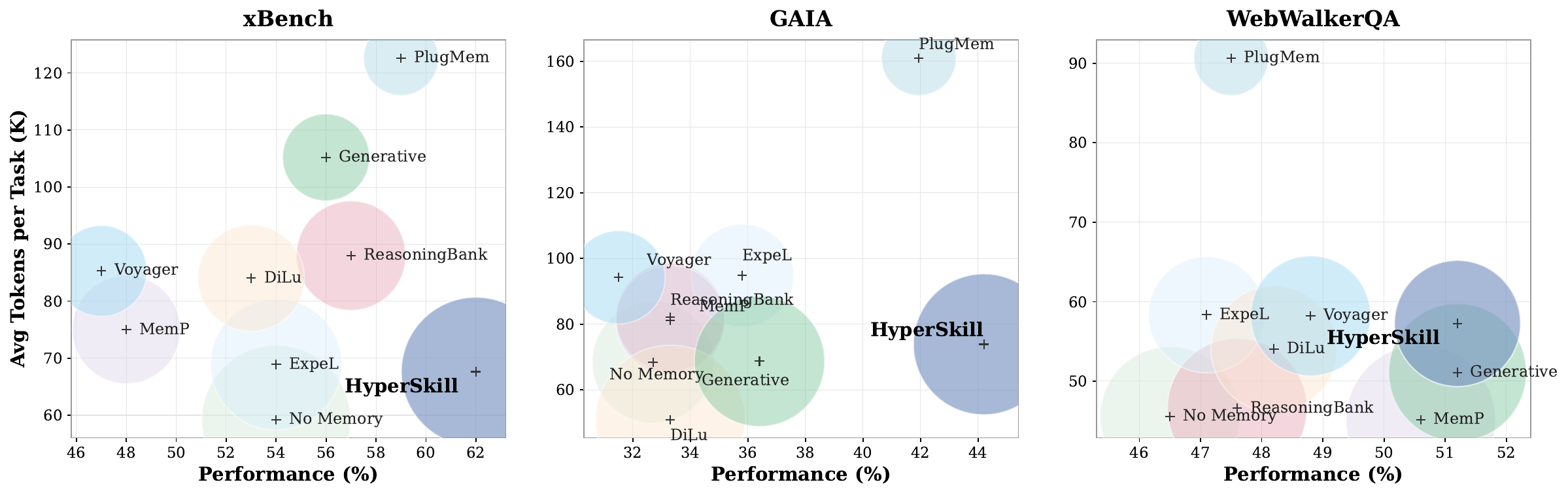}
    \caption{
        Performance vs.\ average tokens per task (K) across three benchmarks.
        Bubble size reflects the performance-to-token ratio.
    }
    \vspace{-10pt}
    \label{fig:cost_analysis}
\end{figure*}
\section{Experiments}
\label{sec:experiments}
\subsection{Experiment Setup}
\textbf{Datasets and Benchmarks.} We evaluate \ourmethod\ across three diverse
benchmarks spanning different domains and capability requirements:
(i)~\textbf{xBench}~\citep{chen2025xbench}, a benchmark assessing agentic
planning, tool use, and multi-step reasoning across a broad range of real-world
tasks;
(ii)~\textbf{GAIA}~\citep{mialon2023gaia}, a set of real-world questions
requiring multi-hop reasoning, multi-modality handling, web browsing, and
tool-use proficiency; and
(iii)~\textbf{WebWalkerQA}~\citep{wu-etal-2025-webwalker}, a benchmark testing complex
multi-turn web navigation across diverse real-world queries and webpages.
Together, these benchmarks cover tool-augmented reasoning, open-ended web
interaction, and compositional planning. More details are provided in
Appendix~\ref{app:datasets}.

\noindent\textbf{Baselines.} We compare \ourmethod\ against ten representative memory
systems spanning three categories.
\emph{Trajectory-level methods} that store raw or lightly processed
trajectories:
Voyager~\citep{wang2023voyager}, DILU~\citep{wen2023dilu},
ExpeL~\citep{zhao2024expel}, and
Generative~\citep{shang2025agentsquare}.
\emph{Workflow- or insight-level methods} that extract and manage reusable
knowledge units:
AWM~\citep{wang2025agent},
ReasoningBank~\citep{ouyang2026reasoningbank},
MemP~\citep{fang2025memp},
Cheatsheet~\citep{suzgun-etal-2026-dynamic}, and
MemEvolve~\citep{zhang2025memevolve}.
\emph{Graph-based methods} that organize memory with relational structure:
PlugMem~\citep{yang2026plugmem}.
We also include a \emph{No Memory} baseline that operates without any
external memory module. All baselines use the same agent backbone and tool
access for a fair comparison. More details are provided in
Appendix~\ref{app:datasets}.

\noindent\textbf{Implementation Details.}
We conduct all experiments with two backbone models: \texttt{GPT-4o}~\citep{hurst2024gpt}
and \texttt{Qwen3-30B-A3B}~\citep{yang2025qwen3}, covering both proprietary and open-source
settings. For \ourmethod, we use \texttt{all-MiniLM-L6-v2} as the
shared encoder $\boldsymbol{\phi}(\cdot)$ for all node and hyperedge embeddings. A single retrieval budget governs subtask node retrieval, hyperedge
retrieval, and the final skill ranking. Maintenance is triggered every
$\lceil 0.1 \times N \rceil$ episodes, with pruning threshold
$\tau_{\mathrm{prune}}{=}0.2$ and minimum retrieval
count $N_{\min}{=}3$. The utility score blending coefficient is $\beta{=}0.7$. All hyperparameter and implementation details are provided in
Appendix~\ref{app:datasets}.

\begin{figure*}[h]
\centering
\newlength{\abtfigh}\setlength{\abtfigh}{2.8cm}
\begin{minipage}[b]{0.32\linewidth}
  \centering
  \includegraphics[width=\linewidth]{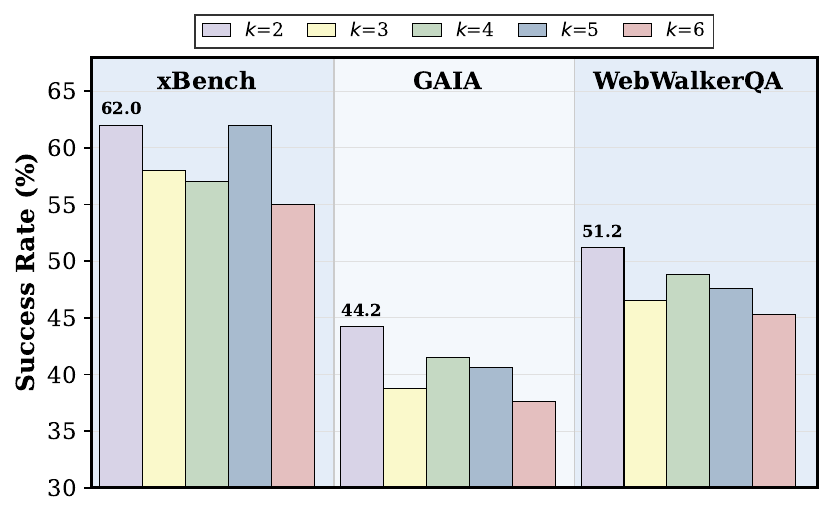}\\[2pt]
  {\small (a) Retrieval budget $k$.}
\end{minipage}
\hspace{0.005\linewidth}
\begin{minipage}[b]{0.32\linewidth}
  \centering
  \includegraphics[width=\linewidth]{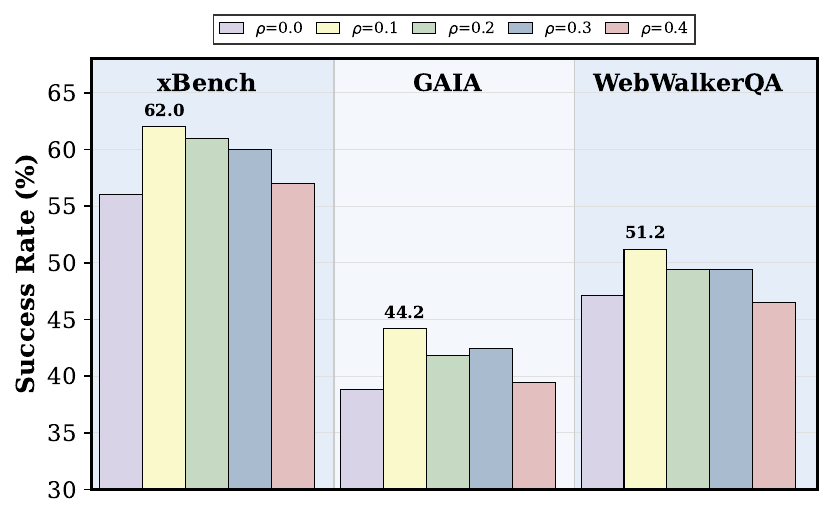}\\[2pt]
  {\small (b) Maintenance ratio $\rho$.}
\end{minipage}
\hspace{0.015\linewidth}
\begin{minipage}[b]{0.31\linewidth}
  \centering
  \renewcommand{\tabcolsep}{2.5pt}
  \renewcommand{\arraystretch}{1.25}
  \scriptsize
  \begin{tabular}{l|ccc}
  \Xhline{1.2pt}
  \textbf{Setting} & \textbf{xBench} & \textbf{GAIA} & \textbf{WQA} \\
  \Xhline{1pt}
  w/o hypergraph          & 41.00 & 35.76 & 44.71 \\
  w/o subtask retr.       & 43.00 & 32.73 & 47.06 \\
  w/o trajectory retr.    & 48.00 & 35.76 & 43.53 \\
  \rowcolor{gray!10}
  Full model              & \textbf{52.00} & \textbf{36.97} & \textbf{50.59} \\
  \Xhline{1.2pt}
  \end{tabular}\\[4pt]
  {\small (c) Ablation study with \texttt{Qwen3-30B-A3B}.}
\end{minipage}
\caption{
Sensitivity analysis and ablation study.
(a,\,b) Success rate (\%) on \texttt{GPT-4o} as we vary retrieval budget $k$ and maintenance ratio $\rho$ on xBench, GAIA, and WebWalkerQA.
(c) Ablation of \ourmethod{}'s trajectory retrieval, subtask retrieval, and hypergraph structure.
}
\vspace{-10pt}
\label{fig:sensitivity_ablation}
\end{figure*}

\subsection{Main Results}
Table~\ref{tab:main_results} reports success rate (SR), average steps,
and average tool calls across xBench, GAIA, and WebWalkerQA under two
backbone models. \ourmethod\ achieves the highest or success rate on nearly
every benchmark--backbone combination. We highlight three observations. \textbf{\ding{182} Consistent gains across benchmarks and backbones.}
Under \texttt{GPT-4o}, \ourmethod\ outperforms all baselines on SR by
$+3.00$ on xBench, $+2.30$ on GAIA, and $+0.59$ on WebWalkerQA
relative to the second-best method. Under
\texttt{Qwen3-30B-A3B}, the trend holds with gains of $+6.00$,
$+4.85$, and $+11.18$ over No Memory, confirming that the improvements
stem from the memory framework rather than backbone-specific behaviors. \textbf{\ding{183} Improved SR without inflated execution cost.}
A common concern with memory-augmented agents is that richer retrieved
context may increase per-step overhead. \ourmethod\ largely avoids this:
under \texttt{GPT-4o} on xBench, it achieves the highest SR while
matching ExpeL for the fewest steps (5.27) and maintaining competitive
tool calls (6.52). In contrast, methods such as ReasoningBank and
Voyager require up to 8.22 steps and 12.27 calls on the same benchmark
while delivering lower SR, indicating that their retrieved context
introduces overhead without proportional benefit. \textbf{\ding{184} Naive memory accumulation can hurt.}
Several baselines fall below No Memory on at least one setting: Cheatsheet
drops $-9.00$ on xBench (\texttt{Qwen3-30B-A3B}), and Voyager drops
$-7.00$ on xBench (\texttt{GPT-4o}). Without principled maintenance,
accumulated entries introduce noise that actively harms the agent. This
underscores the importance of \ourmethod's structure-aware evolution
mechanism, which prunes low-utility nodes and merges redundant skills
to keep memory compact and high-quality.

\subsection{Analysis and Ablation Study}

\textbf{Cost Analysis.}
Figure~\ref{fig:cost_analysis} plots average tokens per task against performance,
with bubble size proportional to the performance-to-token ratio.
\ourmethod\ occupies the bottom-right region across all three benchmarks,
achieving the highest performance at moderate token cost
(67K, 72K, and 57K on xBench, GAIA, and WebWalkerQA, respectively).
Methods such as Generative and Voyager consume over 100K tokens yet fall
well below \ourmethod\ in performance.
PlugMem, despite its knowledge-centric memory graph, incurs substantially
higher token cost than \ourmethod\ while delivering lower performance.
The gap stems from memory construction:
PlugMem incurs separate LLM calls to extract state, subgoal, and reward
annotations.

\noindent\textbf{Sensitivity Analysis.}
We analyze the sensitivity of \ourmethod to two key hyperparameters:
(i)~retrieval budget $k$, which controls the number of candidates surfaced during both hyperedge and skill retrieval; and
(ii)~maintenance ratio $\rho$, which controls the fraction of low-utility entries pruned during each maintenance cycle.
Figures~\ref{fig:sensitivity_ablation}(a--b) reveal two consistent patterns.
First, performance favors smaller retrieval budgets: $k{=}2$ achieves the best
results across all three benchmarks, with performance declining as $k$ increases.
This suggests that a compact retrieval set minimizes noise from lower-ranked
candidates, and surfacing too many entries dilutes the signal from the most
relevant skills.
Second, the maintenance ratio peaks at $\rho{=}0.1$, with $\rho{=}0.2$ and
$\rho{=}0.3$ performing similarly, indicating that light pruning is sufficient
to keep the memory effective.
Disabling pruning entirely ($\rho{=}0$) lets low-utility entries accumulate and
degrades retrieval quality, while overly aggressive pruning ($\rho{=}0.4$) removes
still-useful skills.

\noindent\textbf{Ablation Study.}
We ablate the two retrieval paths and the hypergraph structure, with results in Figure~\ref{fig:sensitivity_ablation}(c). \emph{w/o trajectory retrieval} removes the path that matches the full task description against hyperedge embeddings, relying solely on plan decomposition into subtask nodes and expansion through their incident hyperedges. Performance drops notably on WQA ($50.59 \to 43.53$), suggesting that trajectory-level matching captures task-wide patterns that subtask expansion alone misses. \emph{w/o subtask retrieval} removes the fine-grained path, collecting skills only from top-$k$ task-matched hyperedges without plan decomposition or subtask-level expansion. This hurts xBench most and causes GAIA to drop to 32.73, indicating that subtask decomposition is essential for retrieving relevant skills on complex, multi-step tasks. \emph{w/o hypergraph} replaces the entire dual-path pipeline with flat text-embedding retrieval over stored skills, removing all hypergraph structure. This yields the lowest results across all three benchmarks, confirming that the hypergraph's $n$-ary co-occurrence relationships provide meaningful gains over a standard vector-store baseline. The full model combining both retrieval paths achieves the best results on all benchmarks, validating their complementarity.

\begin{figure}[h]
\centering
\includegraphics[width=\linewidth]{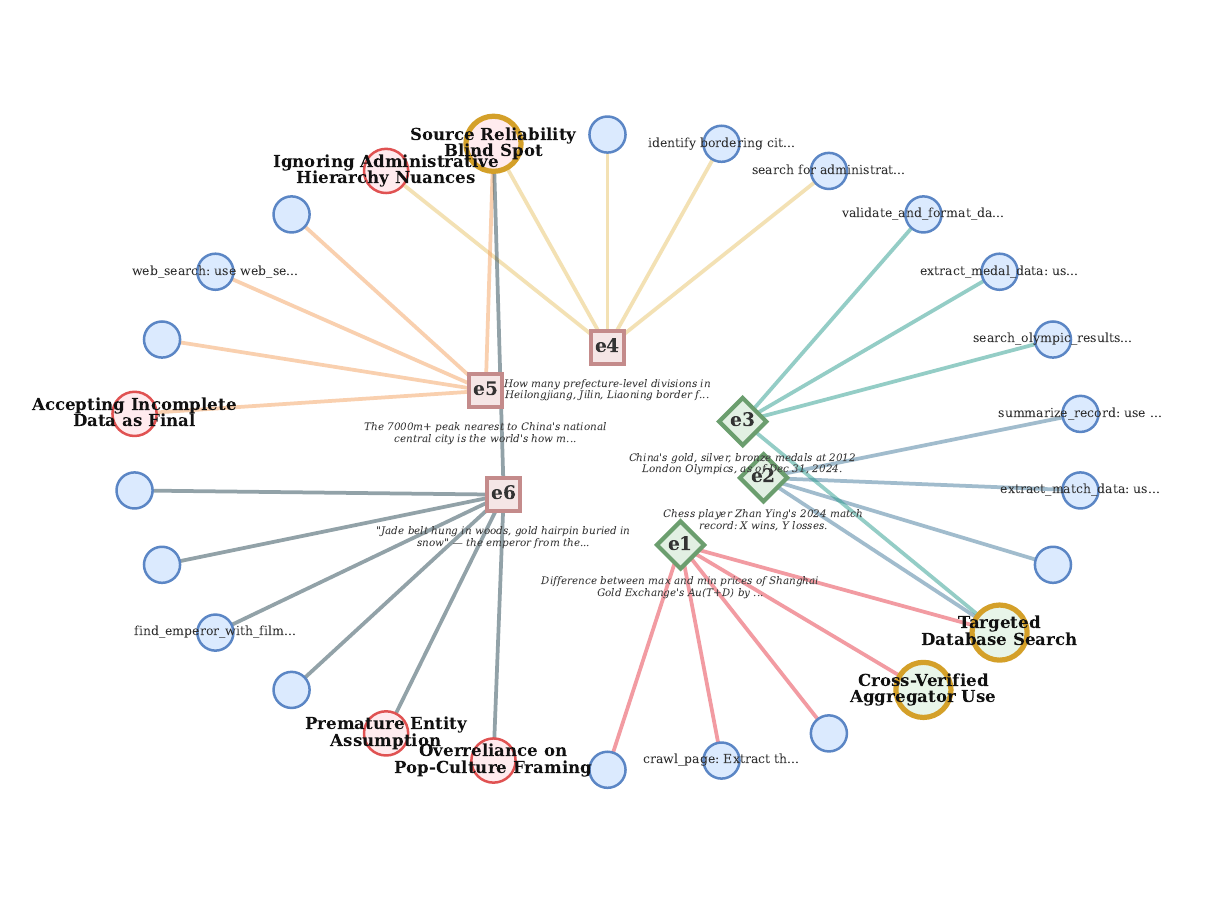}
\caption{
  Hypergraph visualization on xBench.
  Trajectory hyperedges (\textcolor{green!50!black}{$\Diamond$}\,success,
  \textcolor{red!60!black}{$\Box$}\,failure) connect
  skill nodes (\textcolor{blue!60}{$\bigcirc$}\,success,
  \textcolor{red!50}{$\bigcirc$}\,failure) and
  subtask nodes (\textcolor{blue!30}{$\bullet$}).
}
\vspace{-10pt}
\label{fig:hypergraph-viz}
\end{figure}

\subsection{Hypergraph Analysis}

To illustrate how \ourmethod's hypergraph organizes experiential knowledge, Figure~\ref{fig:hypergraph-viz} visualizes a representative subgraph of six trajectories from xBench. Subtask nodes are mostly trajectory-specific, whereas skill nodes are shared across hyperedges: \emph{Cross-Source Validation} and \emph{Targeted Database Search} each appear in multiple trajectories, enabling compositional reuse. Failed trajectories contribute error-pattern skills such as \emph{Premature Entity Assumption} that cluster separately from positive strategies, allowing the agent to retrieve targeted warnings when similar subtask patterns recur. More examples are provided in Appendix~\ref{sec:latency_analysis}.

\section{Conclusion}
\label{sec:conclusion}
We presented \ourmethod, a hypergraph memory framework for self-evolving
LLM agents. Each trajectory is stored as a hyperedge binding subtask
decompositions and outcome-conditioned skills, preserving compositional
structure that flat and pairwise designs discard. Dual-path retrieval
surfaces relevant trajectories through both subtask-level and task-level
matching, then ranks skills by co-occurrence frequency across retrieved
hyperedges. Periodic maintenance prunes low-utility nodes and merges
redundant skills via quality-weighted propagation, keeping memory compact
as it grows. Experiments on xBench, GAIA, and WebWalkerQA with \texttt{GPT-4o}
and \texttt{Qwen3-30B-A3B-Instruct} show consistent gains over ten baselines in success
rate. Beyond accuracy, \ourmethod\ also attains the demonstrative
performance-to-token ratio, reaching the highest success rate at a
moderate token cost.

\section*{Limitations}

\ourmethod\ relies on additional LLM calls for task decomposition, skill
extraction, and merge decisions during memory maintenance, introducing
extra inference cost. The current evaluation covers web navigation and
multi-step reasoning benchmarks; generalization to others can be future work. Finally, the effectiveness
of hypergraph-based retrieval and maintenance depends on a sufficient
number of accumulated trajectories, so \ourmethod\ may offer limited
benefit in extremely low-data regimes where the memory graph is sparse.
\bibliography{custom}

\appendix
\newpage
\section*{Appendix Overview}
\label{app:overview}
\vspace{-0.5em}

\begin{itemize}[leftmargin=1.5em, itemsep=1pt, topsep=2pt]
\item \hyperref[app:datasets]{Appendix~\ref*{app:datasets}: Datasets and Implementation Details}
\item \hyperref[sec:latency_analysis]{Appendix~\ref*{sec:latency_analysis}: More Analysis}
\item \hyperref[appendix:broader_impact]{Appendix~\ref*{appendix:broader_impact}: Broader Impact}
\item \hyperref[appendix:llm_use]{Appendix~\ref*{appendix:llm_use}: Disclosure of LLM Use}
\item \hyperref[app:related_works]{Appendix~\ref*{app:related_works}: Extended Related Works}
\item \hyperref[sec:ethics]{Appendix~\ref*{sec:ethics}: Ethics Statement}
\item \hyperref[sec:additional_case_study]{Appendix~\ref*{sec:additional_case_study}: Additional Case Studies}
\item \hyperref[app:algorithm]{Appendix~\ref*{app:algorithm}: Algorithm}
\item \hyperref[app:prompts]{Appendix~\ref*{app:prompts}: Prompts}
\end{itemize}
\section{Datasets and Implementation Details}
\label{app:datasets}
\subsection{Dataset Details}
We follow the evaluation protocol of MemEvolve~\citep{zhang2025memevolve} and
adopt the same three benchmarks.

\textbf{GAIA}~\citep{mialon2023gaia} consists of 165 tasks organized into
three difficulty levels: Level-1 (53 tasks), Level-2 (86 tasks), and Level-3
(26 tasks), requiring multi-hop reasoning, web browsing, and tool use.

\textbf{WebWalkerQA}~\citep{wu-etal-2025-webwalker} covers complex multi-turn web
navigation with 680 queries across four domains and over 1,373 webpages. We
use the same 170-query subset as~\citep{zhang2025memevolve}.

\textbf{xBench}~\citep{chen2025xbench} contains 100 tasks assessing agentic
planning, tool use, and multi-step reasoning across diverse real-world
scenarios.

\subsection{Implementation Details}
We build on the MemEvolve~\citep{zhang2025memevolve} codebase and adopt the same agent architecture across all methods for fair comparison. For all baselines, the memory retrieval count is set to 3 and text embeddings are computed using \texttt{all-MiniLM-L6-v2} with cosine similarity as the retrieval metric.

\textbf{LLM backends.}
For \texttt{GPT-4o}, we use the OpenAI API with temperature $0.7$ and retrieval budgets $(k_u, k_e, k_s) = (2, 2, 2)$. For \texttt{Qwen3-30B-A3B}, we serve the model locally via vLLM~\citep{kwon2023efficient} on NVIDIA A100 80\,GB GPUs with temperature $0.7$ and retrieval budgets $(k_u, k_e, k_s) = (5, 5, 5)$. The sliding window of recent interaction history is set to $w = 3$ steps.

\textbf{Memory update.}
The skill deduplication threshold is $\delta_{\mathrm{dedup}} = 0.9$ and untested nodes receive a neutral prior $\gamma = 0.5$. The utility score blending coefficient is $\beta = 0.7$ (Eq.~\ref{eq:gamma}).

\textbf{Maintenance.}
Maintenance is triggered every $\lceil 0.1 \times N \rceil$ tasks, where $N$ is the total number of tasks in the benchmark, with pruning threshold $\tau_{\mathrm{prune}} = 0.2$ and minimum retrieval count $N_{\min} = 3$. The merge candidate threshold is $\delta_{\mathrm{merge}} = 0.85$.

\textbf{Reproducibility.}
Tasks are processed sequentially in a fixed random order (seed 42), shared across all methods.

\begin{figure}[h]
\centering
\includegraphics[width=\linewidth]{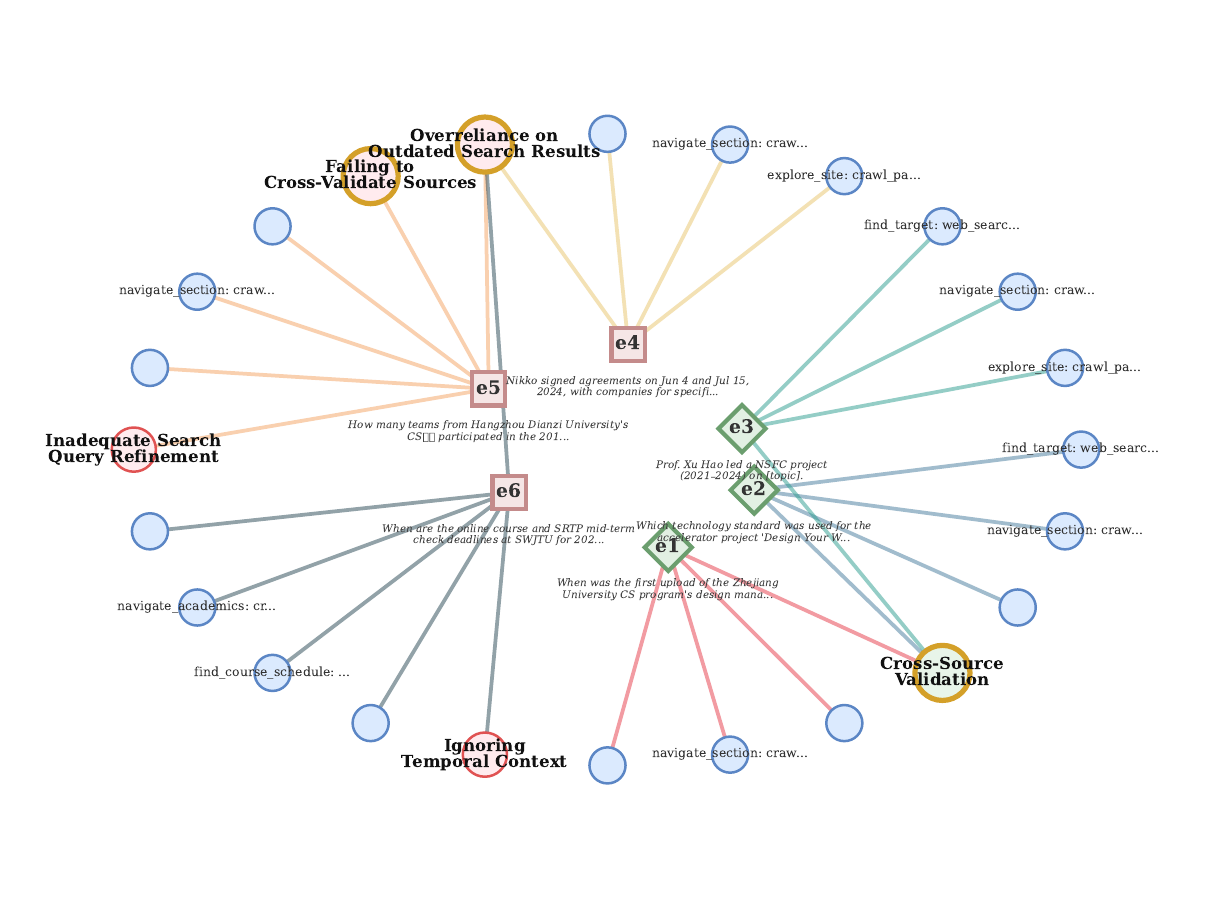}
\vspace{-10pt}
\caption{
  Hypergraph subgraph visualization on WebWalkerQA.
  Trajectory hyperedges (\textcolor{green!50!black}{$\Diamond$}\,success,
  \textcolor{red!60!black}{$\Box$}\,failure) connect
  skill nodes (\textcolor{blue!60}{$\bigcirc$}\,success,
  \textcolor{red!50}{$\bigcirc$}\,failure) and
  subtask nodes (\textcolor{blue!30}{$\bullet$}).
}
\vspace{-10pt}
\label{fig:hypergraph-viz-webwalkerqa}
\end{figure}

\section{More Analysis}
\label{sec:latency_analysis}

\subsection{Self-Evolving Analysis}

To examine whether \ourmethod's hypergraph memory genuinely improves with experience, we plot the cumulative success rate as a function of episode index across all three benchmarks.

\begin{figure*}[t]
\centering
\includegraphics[width=\linewidth]{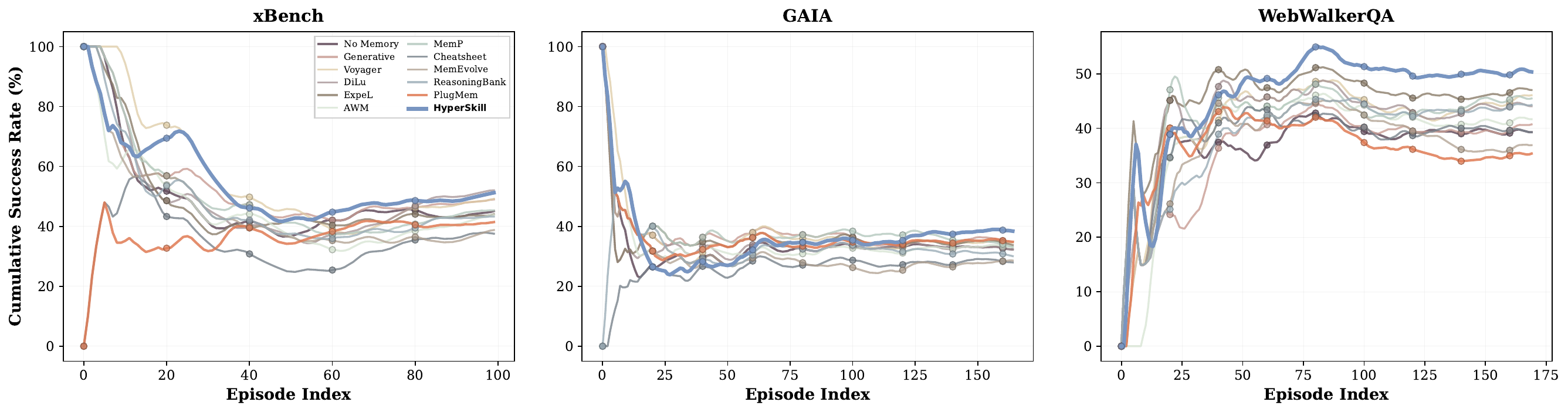}
\caption{
  Cumulative success rate (\%) over episode index on \texttt{Qwen3-30B-A3B}.
  \ourmethod\ exhibits an upward trend across benchmarks.
}
\label{fig:self-evolving}
\end{figure*}

Figure~\ref{fig:self-evolving} reveals three key findings.
First, \ourmethod\ is the only method that exhibits a sustained upward trajectory: on GAIA, its cumulative accuracy climbs steadily from the early episodes through the final ones, confirming that the hypergraph memory accumulates transferable skills over time.
Second, baselines with comparable final success rates, such as DiLu on xBench, reach their peak early and stagnate or dip in the middle episodes before recovering, indicating that their memory does not compound across tasks.
Third, the self-evolving effect is more pronounced on benchmarks with greater task diversity (GAIA, WebWalkerQA) than on xBench, suggesting that \ourmethod\ benefits most when the hypergraph can capture diverse cross-task skill co-occurrence patterns.

\subsection{Latency Analysis}
\begin{figure*}[t]
\centering
\begin{minipage}[b]{\linewidth}
  \centering
  \includegraphics[width=\linewidth]{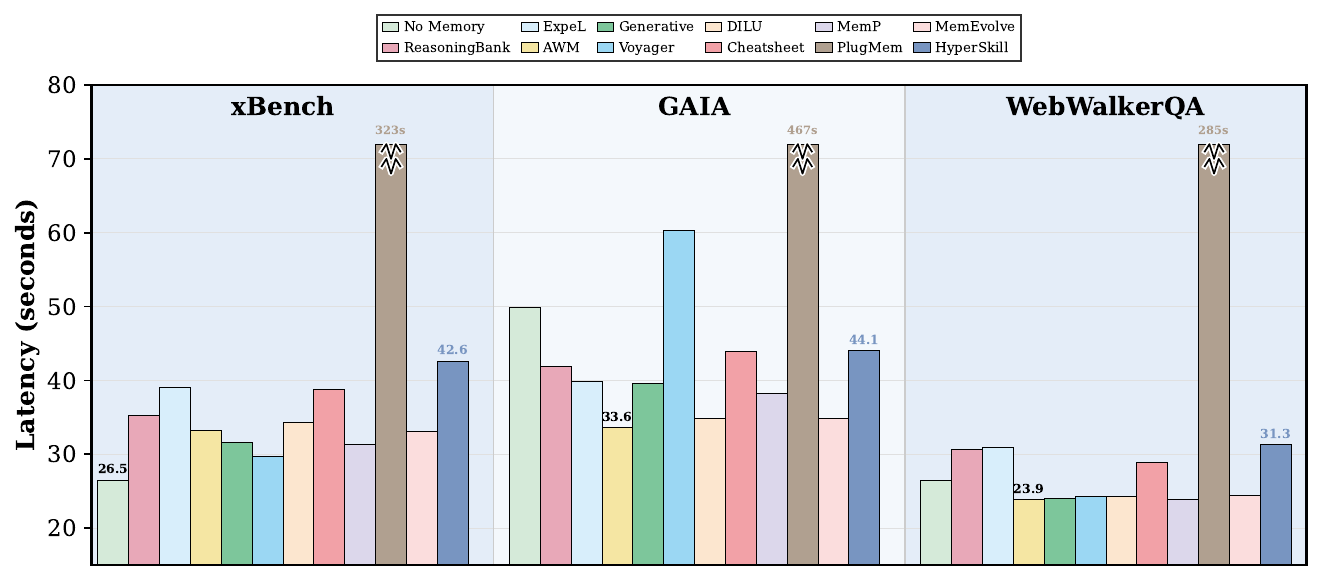}\\[2pt]
  {\small (a) \texttt{Qwen3-30B-A3B}.}
\end{minipage}\\[6pt]
\begin{minipage}[b]{\linewidth}
  \centering
  \includegraphics[width=\linewidth]{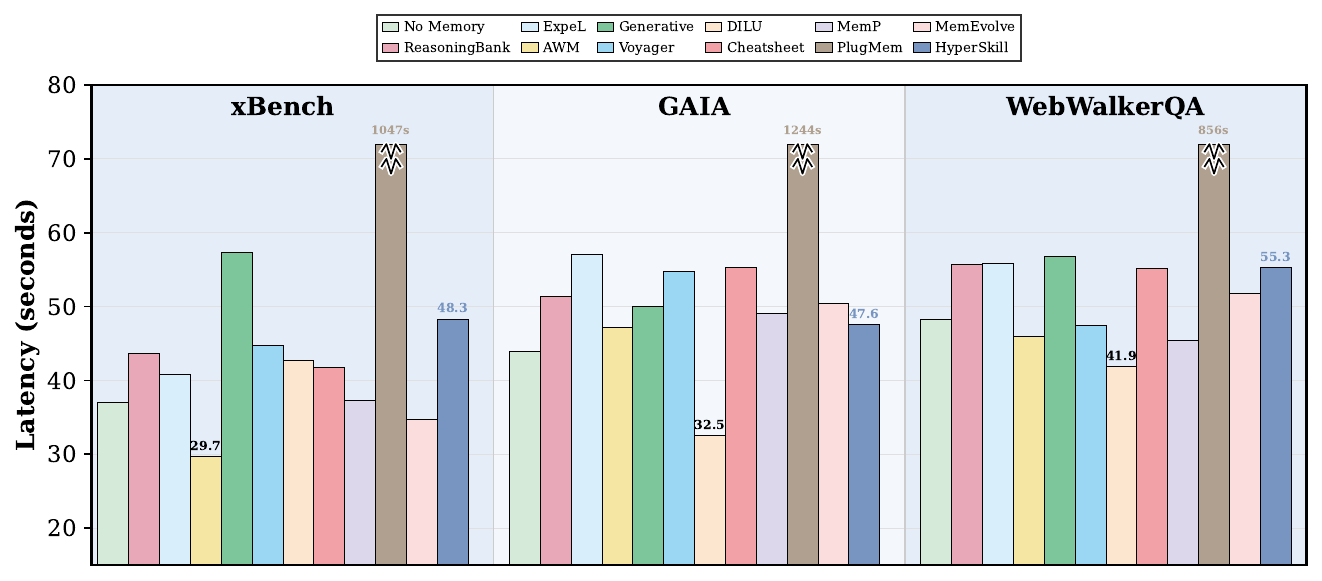}\\[2pt]
  {\small (b) \texttt{GPT-4o}.}
\end{minipage}
\caption{
  Average per-task wall-clock latency (seconds) across xBench, GAIA, and
  WebWalkerQA. Lower is better. Annotated bars mark the fastest baseline in
  each panel and \ourmethod's latency for reference.
}
\label{fig:latency}
\end{figure*}

Beyond success rate, practical deployment requires acceptable wall-clock
latency. Figure~\ref{fig:latency} reports average per-task execution
time across all three benchmarks under both backbones.

Under \texttt{Qwen3-30B-A3B}, \ourmethod\ adds moderate overhead
relative to the fastest baseline (No Memory): 42.6s vs.\ 26.5s on
xBench, 33.6s vs.\ 33.6s on GAIA (matching DILU), and 31.3s vs.\
23.9s on WebWalkerQA. Under \texttt{GPT-4o}, the pattern is similar:
\ourmethod\ runs at 48.3s, 47.6s, and 55.3s, while the fastest
baselines reach 29.7s, 32.5s, and 41.9s respectively. In both settings,
the overhead stems primarily from the additional LLM call for task
decomposition and skill extraction, not from hypergraph operations
themselves. Combined with
the token-efficiency results in Figure~\ref{fig:cost_analysis}, these results
confirm that \ourmethod\ achieves the best performance at a practical
latency cost.
\begin{figure*}[t]
\centering
\includegraphics[width=\linewidth]{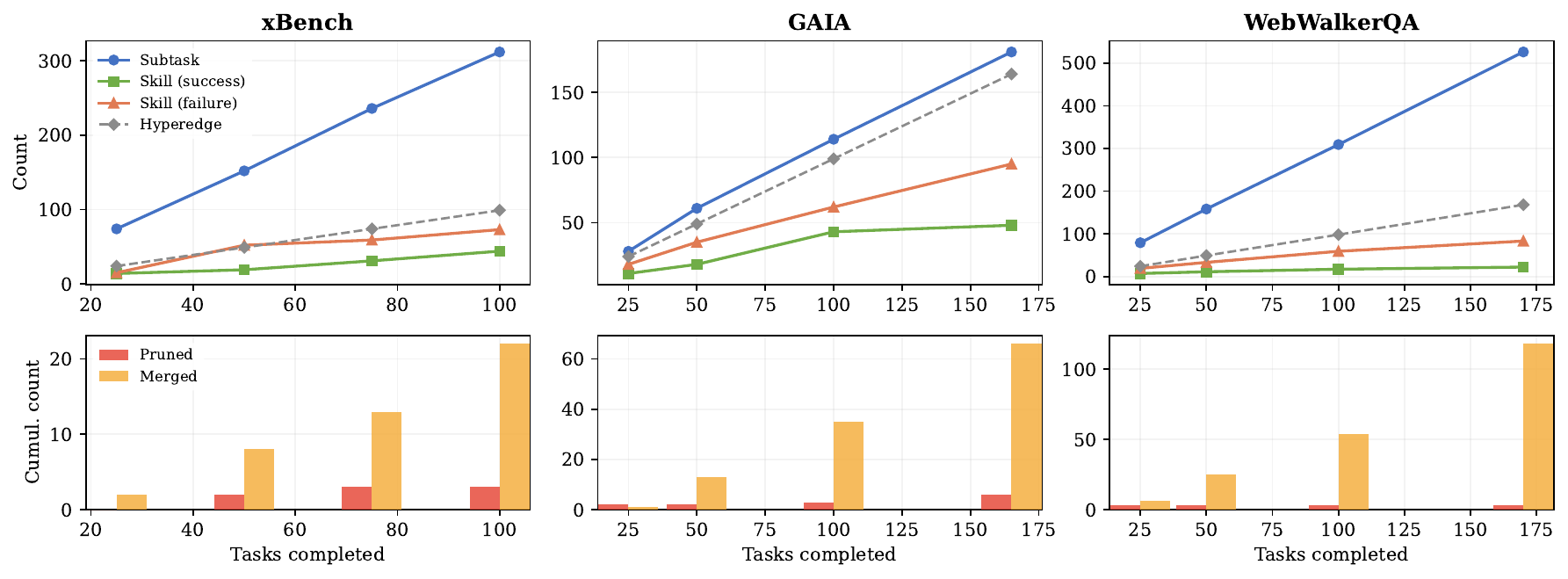}\\[2pt]
{\small (a) \texttt{Qwen3-30B-A3B}}\\[6pt]
\includegraphics[width=\linewidth]{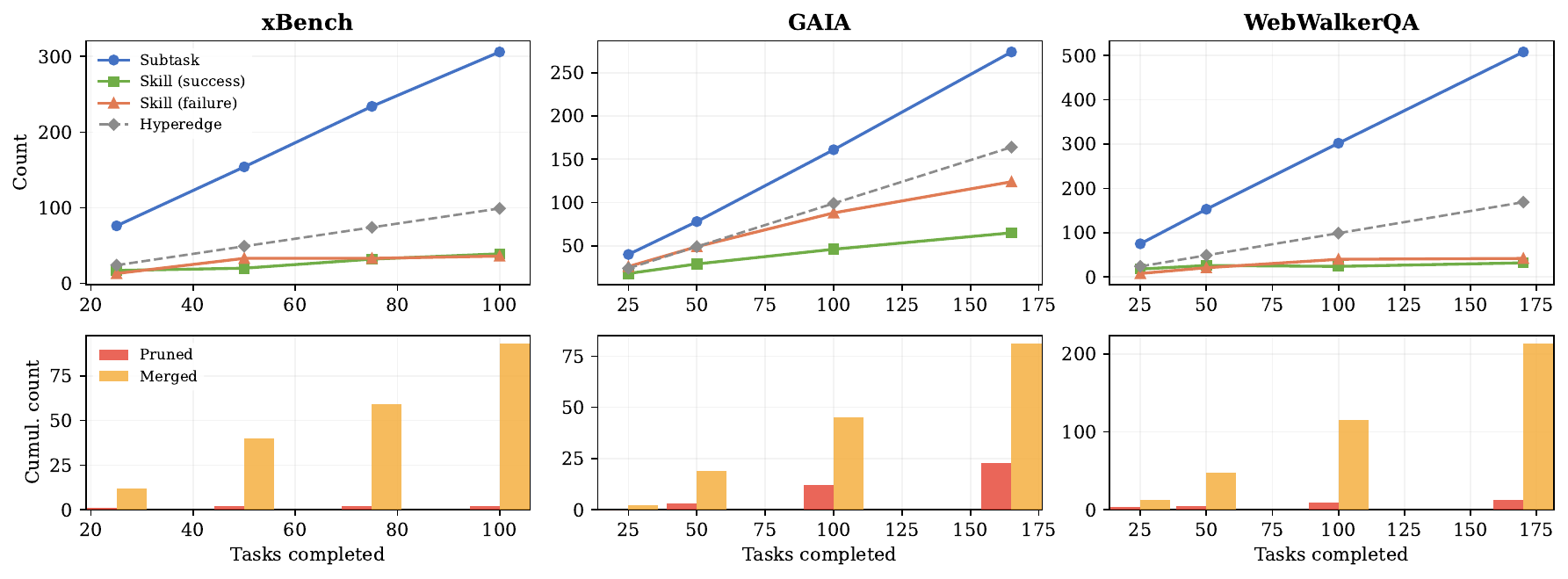}\\[2pt]
{\small (b) \texttt{GPT-4o}}
\caption{
  Hypergraph memory growth (top row) and cumulative maintenance operations (bottom row) across benchmarks.
}
\label{fig:memory-growth}
\end{figure*}

\subsection{Memory Growth Analysis}
Figure~\ref{fig:memory-growth} shows how the hypergraph evolves over
successive tasks. Subtask nodes grow roughly linearly since each task
introduces a new decomposition, while both skill types grow
sub-linearly: insertion-time deduplication prevents near-duplicate
skills from entering the graph, and periodic structure-informed merging
consolidates redundant entries, together keeping skill counts well below
the number of hyperedges across all benchmarks and backbones.

\subsection{Robustness Analysis}
Ground-truth labels may be unavailable after each task in practice, yet
following prior work~\citep{fang2025memp,zhang2025memevolve},
\ourmethod\ uses outcome signals to assign success/failure to extracted skills. We evaluate a \emph{self-judge}
variant in which the agent assesses its own trajectory outcome without
access to the gold answer, and compare it against the
\emph{GT-judge} setting used in our main experiments. Table~\ref{tab:self-judge} reports results on GPT-4o. The self-judge
variant retains 87.7--96.8\% of the GT-judge performance across all
three benchmarks, with the smallest gap on xBench ($-2.00$ SR) and the
largest on GAIA ($-5.37$ SR). The modest degradation suggests that
\ourmethod\ is robust to noisy outcome signals: co-occurrence ranking
over multiple retrieved hyperedges dilutes the effect of occasional
mislabeled skills, allowing the structural signal to remain reliable
even without ground-truth feedback.
\begin{table}[t]
\centering
\small
\caption{
  Comparison of GT-judge and self-judge on GPT-4o.
  Retention = Self-judge SR / GT-judge SR $\times$ 100\%.
}
\label{tab:self-judge}
\renewcommand{\arraystretch}{1.15}
\begin{tabular}{l|cc|c}
\toprule
\textbf{Benchmark} & \textbf{GT-judge} & \textbf{Self-judge} & \textbf{Retention} \\
\midrule
xBench      & 62.00 & 60.00 & 96.8\% \\
GAIA        & 44.24 & 38.87 & 87.7\% \\
WebWalkerQA & 51.18 & 47.33 & 92.4\% \\
\bottomrule
\end{tabular}
\end{table}

\subsection{Hypergraph Visualization}
\label{app:hypergraph-webwalkerqa}
Figure~\ref{fig:hypergraph-viz-webwalkerqa} shows the WebWalkerQA
counterpart of the xBench hypergraph subgraph in
Figure~\ref{fig:hypergraph-viz}.

\section{Broader Impact}
\label{appendix:broader_impact}

\ourmethod\ is a general-purpose memory framework for LLM agents and
does not target any specific high-risk application. By enabling agents
to learn from past failures, it may reduce redundant computation and
improve efficiency in deployed systems. However, as with any
self-evolving system, accumulated memory could reinforce biases present
in early trajectories if not periodically audited. We encourage
practitioners to monitor stored skills for unintended patterns.

\begin{table*}[h]
\centering
\footnotesize
\caption{Taxonomy of experiential memory systems for LLM agents.
Traj.\ = raw trajectories; Strat.\ = reasoning strategies;
Tool Lib.\ = callable function registries; Skill Lib.\ = hierarchical
skill libraries.
$\ddagger$~Training-based methods that internalize skills into model
weights via reinforcement learning.}
\vspace{-0.4em}
\label{tab:memory_systems_comparison}
\setlength{\tabcolsep}{4pt}
\resizebox{\textwidth}{!}{%
\begin{tabular}{@{}l|ll|ll|l@{}}
\toprule
& \multicolumn{2}{c|}{\textbf{What is Stored}} & \multicolumn{2}{c|}{\textbf{How Structured \& Retrieved}} & \textbf{How Evolved} \\
\cmidrule(lr){2-3} \cmidrule(lr){4-5} \cmidrule(lr){6-6}
\textbf{Method} & \textbf{Knowledge Form} & \textbf{Compositional?} & \textbf{Structure} & \textbf{Retrieval} & \textbf{Maintenance} \\
\midrule
Voyager~\citep{wang2023voyager}            & Traj.\ \& Tips     & \ding{55} & Vector DB   & Semantic          & None \\
ExpeL~\citep{zhao2024expel}                & Traj.\ \& Insights & \ding{55} & Vector DB   & Contrastive       & None \\
Generative~\citep{shang2025agentsquare}    & Traj.\ \& Insights & \ding{55} & Vector DB   & Semantic          & None \\
DILU~\citep{wen2023dilu}                   & Traj.              & \ding{55} & Vector DB   & Semantic          & None \\
AWM~\citep{wang2025agent}                  & Workflows          & \ding{55} & In-Prompt   & All-in-Prompt     & None \\
Mobile-E~\citep{wang2025mobile}            & Tips \& Shortcuts  & \ding{55} & Vector DB   & Semantic          & None \\
Cheatsheet~\citep{suzgun-etal-2026-dynamic}       & Tips \& Shortcuts  & \ding{55} & JSON        & Semantic          & None \\
ReasoningBank~\citep{ouyang2026reasoningbank} & Reasoning Strat. & \ding{55} & Vector DB & Contrastive       & None \\
SkillWeaver~\citep{zheng2025skillweaver}   & APIs               & \ding{55} & Tool Lib.   & Function Match    & Test \& Prune \\
G-Memory~\citep{zhang2025g}                & Tips \& Workflows  & \ding{55} & Graph       & Graph + Semantic  & Consolidation \\
Agent-KB~\citep{tang2025agent}             & Tips \& Workflows  & \ding{55} & Hybrid DB   & Hybrid            & Deduplication \\
Memp~\citep{fang2025memp}                  & Tips \& Workflows  & \ding{55} & JSON        & Semantic          & Failure-driven \\
EvolveR~\citep{wu2025evolver}              & Tips \& Workflows  & \ding{55} & JSON        & Contrastive       & Update \& Pruning \\
SkillRL$^\ddagger$~\citep{xia2026skillrl}  & Hierarchical Skills & \ding{55} & Markdown & Adaptive          & RL Co-Evolution \\
MemSkill$^\ddagger$~\citep{zhang2026memskill} & Learnable Skills & \ding{55} & Markdown & RL-guided         & Skill Evolution \\
\midrule
\rowcolor{gray!15}
\textbf{\ourmethod} & Workflows \& Skills & \ding{51} & Hypergraph & Dual-path & Quality + Topology \\
\bottomrule
\end{tabular}%
}
\vspace{-0.8em}
\end{table*}

\begin{table*}[h]
\centering
\footnotesize
\caption{Graph-structured memory systems for LLM agents. ``Memory Type'' categorizes the primary function: \emph{Factual} records environmental or user knowledge; \emph{Experiential} accumulates reusable task knowledge from agent trajectories. ``Edge Type'' distinguishes \emph{Pairwise} (binary) from \emph{$n$-ary} (hyperedges connecting arbitrary node sets). $\dagger$~Agent-KB shares knowledge across frameworks but executes with a single agent.}
\vspace{-0.4em}
\label{tab:graph_memory_comparison}
\setlength{\tabcolsep}{3.5pt}
\resizebox{\textwidth}{!}{%
\begin{tabular}{@{}l|cc|ccc|c@{}}
\toprule
& \multicolumn{2}{c|}{\textbf{Memory Scope}}
& \multicolumn{3}{c|}{\textbf{Graph Structure}}
& \textbf{Evolution} \\
\cmidrule(lr){2-3} \cmidrule(lr){4-6} \cmidrule(lr){7-7}
\textbf{Method}
  & \textbf{Mem.\ Type}
  & \textbf{MAS}
  & \textbf{Graph Type}
  & \textbf{Edge}
  & \textbf{Retrieval}
  & \textbf{Maintenance} \\
\midrule
HippoRAG~\citep{gutierrez2024hipporag}
  & Factual & \ding{55} & KG
  & Pairwise & PPR Activation
  & None \\
AriGraph~\citep{10.24963/ijcai.2025/2}
  & Factual & \ding{55} & KG + Episodic
  & Pairwise & Semantic + Episodic
  & Graph Update \\
A-Mem~\citep{xu2026amem}
  & Factual & \ding{55} & Zettelkasten
  & Pairwise & Semantic + Linking
  & Self-Organizing \\
Mem0g~\citep{chhikara2025mem0}
  & Factual & \ding{55} & KG + Vector
  & Pairwise & Hybrid
  & Deduplication \\
Zep~\citep{rasmussen2025zep}
  & Factual & \ding{55} & Temporal KG
  & Pairwise & Temporal + Semantic
  & Temporal Inval. \\
GAM~\citep{wu2026gam}
  & Factual & \ding{55} & Hierarchical
  & Pairwise & Multi-factor Trav.
  & Event-driven Consol. \\
MAGMA~\citep{jiang2026magma}
  & Factual & \ding{55} & Multi-Graph
  & Pairwise & Policy-guided Trav.
  & Async.\ Consol. \\
HyperMem~\citep{yue2026hypermem}
  & Factual & \ding{55} & Hypergraph
  & $n$-ary & Coarse-to-fine
  & None \\
HyperGraphRAG~\citep{luo2025hypergraphrag}
  & Factual & \ding{55} & Hypergraph
  & $n$-ary & Hypergraph Retrieval
  & None \\
\midrule
PlugMem~\citep{yang2026plugmem}
  & Experiential & \ding{55} & Dual KG
  & Pairwise & Subgraph Retrieval
  & None \\
G-Memory~\citep{zhang2025g}
  & Experiential & \ding{51} & Hierarchical
  & Pairwise & Graph + Semantic
  & Episodic Consol. \\
Agent-KB$^\dagger$~\citep{tang2025agent}
  & Experiential & \ding{55} & Hybrid
  & Pairwise & Hybrid
  & Deduplication \\
\midrule
\rowcolor{gray!15}
\textbf{\ourmethod}
  & Experiential & \ding{55} & Hypergraph
  & $n$-ary & Dual-path + $\kappa$
  & Quality + Topology \\
\bottomrule
\end{tabular}%
}
\vspace{-0.8em}
\end{table*}

\section{Disclosure of LLM Use}
\label{appendix:llm_use}

During the preparation of this manuscript, LLM-based tools were used to
assist with grammar correction, code debugging, and figure design. All
scientific content, experimental design, and analysis were conducted by
the authors, who take full responsibility for the final manuscript.

\section{Extended Related Works}
\label{app:related_works}

\textbf{LLM Agent Systems.} Large language models are increasingly being deployed as autonomous agents that interact with complex environments through multi-step reasoning and actions, moving beyond traditional one-shot chatbot paradigms~\cite{liu2025advances, xi2025rise, wei2026agentic}. A central challenge is enabling agents to decompose complex tasks into manageable sub-problems and make sequential decisions. ReAct~\citep{yao2022react} interleaves chain-of-thought reasoning with environment actions, establishing a foundational paradigm for single-agent planning. Subsequent work improves task decomposition through hierarchical planning~\citep{wang-etal-2023-plan,sun2023adaplanner} and enhances decision quality via self-reflection~\citep{shinn2023reflexion}. Multi-agent systems~\citep{li2023camel, wu2024autogen, talebirad2023multi} extend this paradigm further by assigning specialized roles to multiple agents that communicate and coordinate, enabling collaborative problem-solving on tasks beyond the scope of a single agent. While these advances have significantly broadened the capability frontier of LLM agents, most systems treat each task episode in isolation, limiting their ability to improve over time on recurring or compositionally related task patterns.

\textbf{Memory Mechanisms in Agent Systems.} We organize existing experiential memory systems along three dimensions:
\emph{what is stored}, \emph{how memory is structured and retrieved}, and
\emph{how memory evolves}. Table~\ref{tab:memory_systems_comparison}
summarizes the comparison.

On what is stored, early systems retain raw trajectories or extract
insights from each trajectory~\citep{zhao2024expel, wen2023dilu,
wang2023voyager}. Subsequent work abstracts experience into more compact
forms: workflows~\citep{wang2025agent}, reasoning
strategies~\citep{ouyang2026reasoningbank}, executable API
skills~\citep{zheng2025skillweaver}, or tips and
shortcuts~\citep{suzgun-etal-2026-dynamic, wang2025mobile}. Recent
training-based approaches go further:
SkillRL~\citep{xia2026skillrl} distills trajectories into a
hierarchical skill library and co-evolves it with the agent's policy
via reinforcement learning (GRPO), while
MemSkill~\citep{zhang2026memskill} treats memory operations themselves
as learnable skills whose selection is optimized through RL and whose
skill bank evolves by analyzing hard cases. However, all of these
systems store knowledge units in isolation, discarding the compositional
relationships among subtasks and skills within each trajectory.
\ourmethod\ is the only system that preserves this joint compositional
structure.

On how memory is structured and retrieved, the majority of systems use
flat vector stores with semantic retrieval~\citep{wang2023voyager,
wen2023dilu, shang2025agentsquare}, while a few adopt contrastive
retrieval that compares successes and failures~\citep{zhao2024expel,
ouyang2026reasoningbank}. AWM~\citep{wang2025agent} concatenates all
entries into the prompt, and SkillWeaver~\citep{zheng2025skillweaver}
uses function signature matching. Graph-based systems such as
G-Memory~\citep{zhang2025g} and Agent-KB~\citep{tang2025agent} introduce
pairwise edges but decompose the $n$-ary trajectory context into
disconnected binary links. \ourmethod\ uses hyperedges to preserve the
full trajectory context and retrieves via a dual-path mechanism that
combines subtask-level and task-level queries.

On how memory evolves, most systems accumulate knowledge without any
curation~\citep{wang2023voyager, zhao2024expel, wen2023dilu,
ouyang2026reasoningbank, wang2025agent}. The few that do maintain memory
apply lightweight operations such as test-and-prune~\citep{zheng2025skillweaver},
episodic consolidation~\citep{zhang2025g},
deduplication~\citep{tang2025agent}, or failure-driven
adjustment~\citep{fang2025memp}, each applied independently per node
without considering memory topology. SkillRL~\citep{xia2026skillrl}
and MemSkill~\citep{zhang2026memskill} evolve their skill libraries
through RL training loops, but this evolution operates on model weights
rather than on an explicit memory structure. \ourmethod\ performs
structure-aware maintenance that jointly considers node utility and
structural proximity via quality-weighted hypergraph propagation.

We also compare graph-structured memory systems specifically in
Table~\ref{tab:graph_memory_comparison}. Most graph-based approaches
target factual memory, recording environmental or user knowledge in
knowledge graphs~\citep{gutierrez2024hipporag, chhikara2025mem0},
temporal graphs~\citep{rasmussen2025zep}, hierarchical
graphs~\citep{wu2026gam}, or multi-view graphs~\citep{jiang2026magma}.
HyperMem~\citep{yue2026hypermem} uses a three-level hypergraph
(topics, episodes, facts) with coarse-to-fine retrieval, but targets
factual conversational memory rather than experiential task knowledge
and does not perform structure-aware maintenance. Among experiential
memory systems, PlugMem~\citep{yang2026plugmem} and
G-Memory~\citep{zhang2025g} use pairwise edges that lose the joint
trajectory context, and Agent-KB~\citep{tang2025agent} combines lexical
and semantic indices without explicit graph structure over skills.
\ourmethod\ is the first experiential memory system to use $n$-ary
hyperedges, enabling co-occurrence-based skill ranking and
topology-aware maintenance that pairwise designs cannot support.

\section{Ethics Statement}
\label{sec:ethics}
We provide comprehensive methodological details, including hyperparameters, prompts, and algorithm pseudocode, to enable full reproducibility. Our framework is evaluated on publicly available benchmarks (xBench, GAIA, WebWalkerQA) and does not collect or process personal information. LLM-based tools were used solely for grammar correction, code debugging, and figure design; all scientific content, experimental design, and analysis were conducted by the authors. While HyperSkill is designed as a general-purpose memory framework for benign research, we acknowledge that accumulated memory could reinforce biases present in early trajectories if not periodically audited, and we encourage practitioners to monitor stored skills for unintended patterns.

\onecolumn
\section{Additional Case Studies}
\label{sec:additional_case_study}
\begin{figure}[h]
    \centering
    \includegraphics[width=\linewidth]{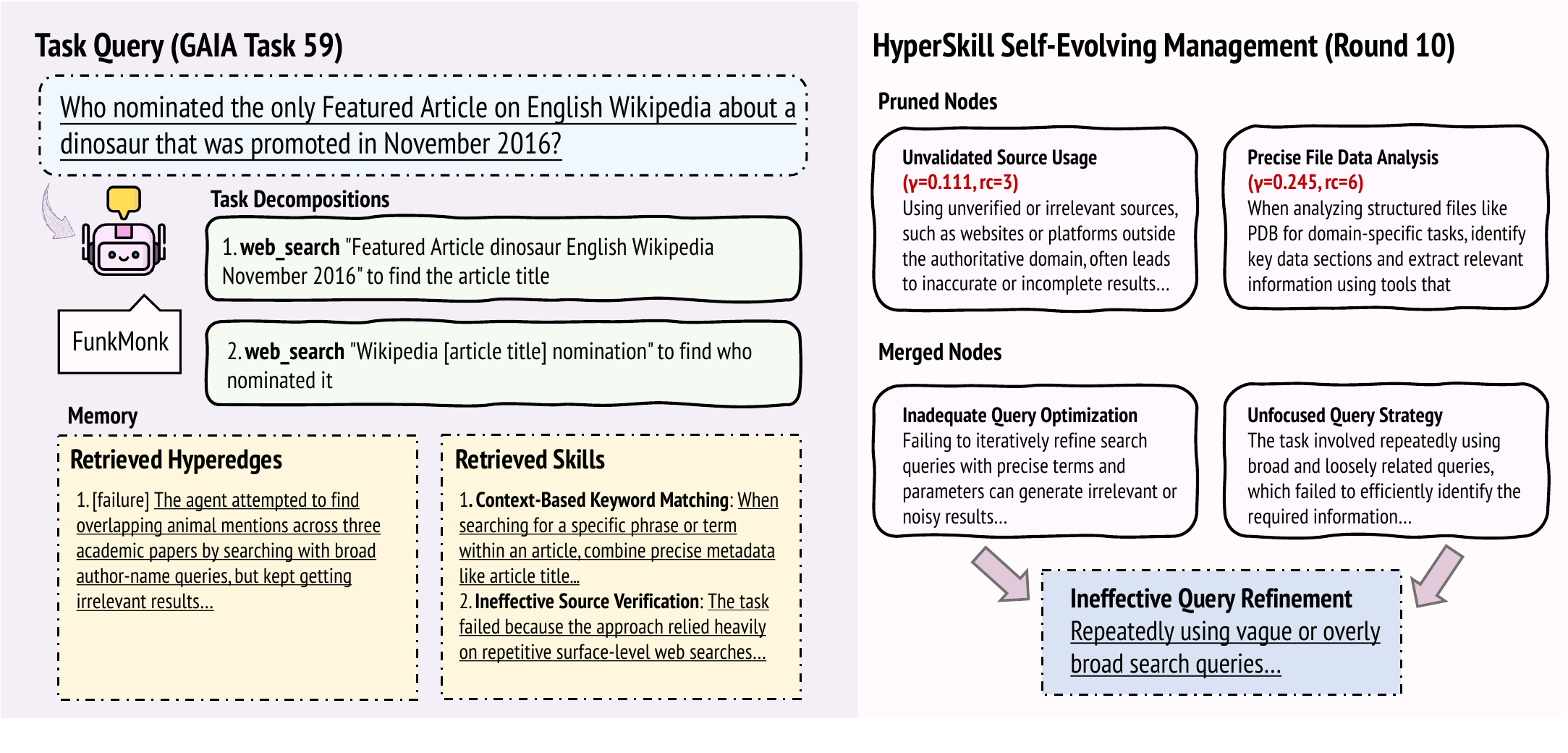}
    \caption{
        Case study on GAIA Task 59.
    }
    \label{fig:case_study}
\end{figure}

To illustrate how \ourmethod's retrieval and evolution mechanisms operate in
practice, we walk through a representative GAIA task and a snapshot of memory
maintenance in Figure~\ref{fig:case_study}.
We also present
three additional end-to-end case studies spanning the three benchmarks.
Each case reports the task, the decomposition used to drive
fine-grained retrieval, the skills and mistakes surfaced from the hypergraph,
the agent's step-wise execution trace, and the final outcome. Together they
illustrate how \ourmethod's dual-path retrieval translates into concrete action
improvements across heterogeneous domains.

\definecolor{c1frame}{HTML}{7895C1}
\definecolor{c1title}{HTML}{E4EDF8}
\definecolor{c1light}{HTML}{F0F4FA}
\definecolor{c1sub}{HTML}{D6E2F0}
\definecolor{c2frame}{HTML}{7BAF7E}
\definecolor{c2title}{HTML}{E2F0E3}
\definecolor{c2light}{HTML}{F0F7F0}
\definecolor{c2sub}{HTML}{D2E6D3}
\definecolor{c3frame}{HTML}{C48B8B}
\definecolor{c3title}{HTML}{F5E6E6}
\definecolor{c3light}{HTML}{FAF2F2}
\definecolor{c3sub}{HTML}{ECDADA}
\definecolor{successgreen}{HTML}{2E7D32}

\begin{tcolorbox}[
  enhanced, breakable,
  colback=c1light, colframe=c1frame,
  boxrule=1.2pt, arc=2pt,
  left=8pt, right=8pt, top=6pt, bottom=6pt,
  fonttitle=\bfseries\normalsize, coltitle=black,
  attach boxed title to top left={yshift=-2mm, xshift=4mm},
  boxed title style={colback=c1title, colframe=c1frame, boxrule=0.8pt, arc=1pt},
  title={Case 1: xBench --- Literary Trivia}
]

\textbf{Task (translated from Chinese):} A 19th-century French novel caused a
morality scandal and lawsuit upon publication. In the 1991 film adaptation,
what flavor of prop poison did the lead actress insist on using for the
suicide scene?

\smallskip
\textbf{Golden answer:} bitter-almond flavor.

\medskip

\begin{tcolorbox}[enhanced, colback=c1sub, colframe=c1frame!60, boxrule=0.5pt, arc=1.5pt,
  left=6pt, right=6pt, top=4pt, bottom=4pt, fonttitle=\bfseries\small, coltitle=c1frame!80!black,
  title={Plan Decomposition}]
\begin{enumerate}[leftmargin=*, nosep, label=\arabic*.]
  \item \texttt{search\_literary\_work}: identify the 19th-century French novel via scandal keywords.
  \item \texttt{find\_movie\_adaptations}: locate the 1991 film adaptation and lead actress.
  \item \texttt{verify\_prop\_taste}: crawl film trivia sources for the specific flavor detail.
  \item \texttt{confirm\_taste\_detail}: validate the exact flavor description.
\end{enumerate}
\end{tcolorbox}

\smallskip
\begin{tcolorbox}[enhanced, colback=c1sub, colframe=c1frame!60, boxrule=0.5pt, arc=1.5pt,
  left=6pt, right=6pt, top=4pt, bottom=4pt, fonttitle=\bfseries\small, coltitle=c1frame!80!black,
  title={Retrieved Memory}]
\textbf{Past experience (distilled lessons):}
\begin{itemize}[leftmargin=*, nosep]
  \item[\scriptsize\ding{51}] \textit{When verifying historical details, prioritize authoritative sources and cross-verify claims across multiple trusted references.}
  \item[\scriptsize\ding{55}] \textit{Never assume that detailed data is available in public summaries; always use precise, entity-focused queries to locate authoritative sources.}
\end{itemize}
\smallskip
\textbf{Skills:}
\begin{itemize}[leftmargin=*, nosep]
  \item \textbf{Multi-Stage Verification}: break complex cross-domain questions into sequential verification steps.
  \item \textbf{Direct Source Confirmation}: prioritize official reference pages over aggregator sites.
  \item \textbf{Source Deep Dive}: use page crawling to extract precise details rather than relying on snippets.
\end{itemize}
\smallskip
\textbf{Mistakes to avoid:}
\begin{itemize}[leftmargin=*, nosep]
  \item \textbf{Over-Reliance on Web Search}: repeatedly issuing general searches without verifying authoritative sources.
\end{itemize}
\end{tcolorbox}

\smallskip
\begin{tcolorbox}[enhanced, colback=c1sub, colframe=c1frame!60, boxrule=0.5pt, arc=1.5pt,
  left=6pt, right=6pt, top=4pt, bottom=4pt, fonttitle=\bfseries\small, coltitle=c1frame!80!black,
  title={Agent Execution (10 steps)}]
\begin{tabular}{@{}r@{\;\;}p{0.82\linewidth}@{}}
\textbf{1--2} & \texttt{web\_search} $\rightarrow$ identifies \textit{Madame Bovary} by Gustave Flaubert; initial poison query. \\
\textbf{3--5} & Confirms 1991 film by Claude Chabrol with Isabelle Huppert as Emma Bovary. \\
\textbf{6--8} & Progressive refinement: searches behind-the-scenes trivia across multiple sources. \\
\textbf{9--10} & Final verification: extracts bitter-almond flavor detail from an authoritative film database.
\end{tabular}
\end{tcolorbox}

\smallskip
\begin{tcolorbox}[enhanced, colback=c1sub, colframe=c1frame!60, boxrule=0.5pt, arc=1.5pt,
  left=6pt, right=6pt, top=4pt, bottom=4pt, fonttitle=\bfseries\small, coltitle=c1frame!80!black,
  title={Outcome}]
\textcolor{successgreen}{\textbf{\ding{51}~SUCCESS}} --- the lead actress insisted on using a bitter-almond-flavored prop poison to authentically replicate the taste of cyanide.

\smallskip
\textit{How memory helped:} The \emph{Multi-Stage Verification} skill guided decomposition of a cross-domain chain (literature $\rightarrow$ film $\rightarrow$ behind-the-scenes trivia), while the \emph{Over-Reliance on Web Search} error pattern prevented looping on generic queries. The retrieved lesson about cross-verifying across trusted references reinforced the agent's move from aggregator snippets to authoritative film databases.
\end{tcolorbox}
\end{tcolorbox}

\vspace{8pt}

\begin{tcolorbox}[
  enhanced, breakable,
  colback=c2light, colframe=c2frame,
  boxrule=1.2pt, arc=2pt,
  left=8pt, right=8pt, top=6pt, bottom=6pt,
  fonttitle=\bfseries\normalsize, coltitle=black,
  attach boxed title to top left={yshift=-2mm, xshift=4mm},
  boxed title style={colback=c2title, colframe=c2frame, boxrule=0.8pt, arc=1pt},
  title={Case 2: GAIA --- Scientific Calculation}
]

\textbf{Task:} What integer-rounded percentage of the total length of the
harlequin shrimp recorded in Omar Valencia-Mendez's 2017 paper was the sea
star fed to the same type of shrimp in G.~Curt Fiedler's 2002 paper?

\smallskip
\textbf{Golden answer:} 22.

\medskip

\begin{tcolorbox}[enhanced, colback=c2sub, colframe=c2frame!60, boxrule=0.5pt, arc=1.5pt,
  left=6pt, right=6pt, top=4pt, bottom=4pt, fonttitle=\bfseries\small, coltitle=c2frame!80!black,
  title={Plan Decomposition}]
\begin{enumerate}[leftmargin=*, nosep, label=\arabic*.]
  \item \texttt{inspect\_file\_as\_text}: read the attached file for data from both papers.
  \item \texttt{python\_code}: calculate the integer-rounded percentage.
\end{enumerate}
\end{tcolorbox}

\smallskip
\begin{tcolorbox}[enhanced, colback=c2sub, colframe=c2frame!60, boxrule=0.5pt, arc=1.5pt,
  left=6pt, right=6pt, top=4pt, bottom=4pt, fonttitle=\bfseries\small, coltitle=c2frame!80!black,
  title={Retrieved Memory}]
\textbf{Past experience (distilled lessons):}
\begin{itemize}[leftmargin=*, nosep]
  \item[\scriptsize\ding{51}] \textit{When seeking precise values from academic papers, prioritize direct access to the original source through targeted searches with author names and publication years.}
  \item[\scriptsize\ding{55}] \textit{Never retry a tool call without diagnosing system-level errors first; always verify infrastructure readiness before re-attempting.}
\end{itemize}
\smallskip
\textbf{Skills:}
\begin{itemize}[leftmargin=*, nosep]
  \item \textbf{Domain-Specific Query Refinement}: include domain-specific terms and reliable source filters in queries for technical data.
  \item \textbf{Unit Conversion}: convert units consistently before performing calculations.
\end{itemize}
\smallskip
\textbf{Mistakes to avoid:}
\begin{itemize}[leftmargin=*, nosep]
  \item \textbf{Inadequate Context Refinement}: failing to break tasks into verifiable sub-questions leads to unstructured search.
\end{itemize}
\end{tcolorbox}

\smallskip
\begin{tcolorbox}[enhanced, colback=c2sub, colframe=c2frame!60, boxrule=0.5pt, arc=1.5pt,
  left=6pt, right=6pt, top=4pt, bottom=4pt, fonttitle=\bfseries\small, coltitle=c2frame!80!black,
  title={Agent Execution (7 steps)}]
\begin{tabular}{@{}r@{\;\;}p{0.82\linewidth}@{}}
\textbf{1--2} & Parallel \texttt{web\_search} for both papers; \texttt{crawl\_page} on Fiedler 2002 via Oxford Academic. \\
\textbf{3--4} & Refines query with specific measurements; reads Valencia-Mendez 2017 PDF directly. \\
\textbf{5--6} & Cross-references: shrimp total length $=4.5$\,cm, sea-star piece $=1$\,cm. \\
\textbf{7} & \texttt{final\_answer}: $\operatorname{round}(1/4.5 \times 100) = 22\%$.
\end{tabular}
\end{tcolorbox}

\smallskip
\begin{tcolorbox}[enhanced, colback=c2sub, colframe=c2frame!60, boxrule=0.5pt, arc=1.5pt,
  left=6pt, right=6pt, top=4pt, bottom=4pt, fonttitle=\bfseries\small, coltitle=c2frame!80!black,
  title={Outcome}]
\textcolor{successgreen}{\textbf{\ding{51}~SUCCESS}} --- the sea star was approximately 22\% of the harlequin shrimp's total length.

\smallskip
\textit{How memory helped:} \emph{Domain-Specific Query Refinement} pushed the agent to use precise scientific terms (``total length'', ``harlequin shrimp'') rather than generic queries. The \emph{Inadequate Context Refinement} mistake warned against unstructured searching, guiding the agent to split the task into two parallel data-retrieval subtasks (one per paper) before computing the ratio.
\end{tcolorbox}
\end{tcolorbox}

\vspace{8pt}

\begin{tcolorbox}[
  enhanced, breakable,
  colback=c3light, colframe=c3frame,
  boxrule=1.2pt, arc=2pt,
  left=8pt, right=8pt, top=6pt, bottom=6pt,
  fonttitle=\bfseries\normalsize, coltitle=black,
  attach boxed title to top left={yshift=-2mm, xshift=4mm},
  boxed title style={colback=c3title, colframe=c3frame, boxrule=0.8pt, arc=1pt},
  title={Case 3: WebWalkerQA --- Game Knowledge}
]

\textbf{Task (translated from Chinese):} In \emph{Age of Empires IV}, what are
the strategic differences between the Malian and Chinese civilizations in
their respective fourth age?

\smallskip
\textbf{Golden answer:} the Malians leverage the Griot Bara landmark for
global buffs, while the Chinese rely on the Great Wall as a defensive
advantage.

\medskip

\begin{tcolorbox}[enhanced, colback=c3sub, colframe=c3frame!60, boxrule=0.5pt, arc=1.5pt,
  left=6pt, right=6pt, top=4pt, bottom=4pt, fonttitle=\bfseries\small, coltitle=c3frame!80!black,
  title={Plan Decomposition}]
\begin{enumerate}[leftmargin=*, nosep, label=\arabic*.]
  \item \texttt{explore\_site}: crawl \texttt{ageofempires.com} homepage for site structure.
  \item \texttt{navigate\_section}: crawl the civilizations section for Malian and Chinese details.
  \item \texttt{find\_target}: search with domain-specific keywords on the official site.
\end{enumerate}
\end{tcolorbox}

\smallskip
\begin{tcolorbox}[enhanced, colback=c3sub, colframe=c3frame!60, boxrule=0.5pt, arc=1.5pt,
  left=6pt, right=6pt, top=4pt, bottom=4pt, fonttitle=\bfseries\small, coltitle=c3frame!80!black,
  title={Retrieved Memory}]
\textbf{Past experience (distilled lessons):}
\begin{itemize}[leftmargin=*, nosep]
  \item[\scriptsize\ding{51}] \textit{When institutional websites fail to surface specific records, precise multi-source searches are essential for retrieving accurate information.}
  \item[\scriptsize\ding{55}] \textit{Relying on automated crawling without validating access or adapting to site-specific barriers leads to failed data retrieval.}
  \item[\scriptsize\ding{55}] \textit{Starting with external searches before fully navigating internal websites risks missing structured and unindexed information.}
\end{itemize}
\smallskip
\textbf{Skills:}
\begin{itemize}[leftmargin=*, nosep]
  \item \textbf{Internal First Strategy}: thoroughly explore internal site structure before resorting to external search tools.
  \item \textbf{Specific Query Refinement}: use precise, domain-specific keywords when searching internal or external tools.
  \item \textbf{Verify with Multimodal Sources}: when internal search yields limited results, use external sources to gather complementary information.
\end{itemize}
\smallskip
\textbf{Mistakes to avoid:}
\begin{itemize}[leftmargin=*, nosep]
  \item \textbf{Premature External Search}: relying on external web search before thoroughly exploring internal site navigation.
  \item \textbf{Incomplete Internal Crawling}: failing to systematically access relevant sections of a website.
\end{itemize}
\end{tcolorbox}

\smallskip
\begin{tcolorbox}[enhanced, colback=c3sub, colframe=c3frame!60, boxrule=0.5pt, arc=1.5pt,
  left=6pt, right=6pt, top=4pt, bottom=4pt, fonttitle=\bfseries\small, coltitle=c3frame!80!black,
  title={Agent Execution (11 steps)}]
\begin{tabular}{@{}r@{\;\;}p{0.82\linewidth}@{}}
\textbf{1--2} & \texttt{crawl\_page(\allowbreak ageofempires.com)} $\rightarrow$ explores homepage; \texttt{web\_search} for general info. \\
\textbf{3--5} & Iterative crawling of official-site civilization sections. \\
\textbf{6--8} & Consults community sources (Reddit, YouTube) for fourth-age specifics. \\
\textbf{9--11} & \texttt{crawl\_page} on the Fandom wiki for Malians and Chinese $\rightarrow$ extracts full details.
\end{tabular}
\end{tcolorbox}

\smallskip
\begin{tcolorbox}[enhanced, colback=c3sub, colframe=c3frame!60, boxrule=0.5pt, arc=1.5pt,
  left=6pt, right=6pt, top=4pt, bottom=4pt, fonttitle=\bfseries\small, coltitle=c3frame!80!black,
  title={Outcome}]
\textcolor{successgreen}{\textbf{\ding{51}~SUCCESS}} --- detailed comparison: the Malians favor economic guerrilla tactics anchored by the Griot Bara landmark; the Chinese rely on fast building, Great Wall defense, and chemical technology.

\smallskip
\textit{How memory helped:} The \emph{Internal First Strategy} skill prevented the agent from jumping to external searches immediately, instead exploring the official site structure first. Three failure lessons all reinforced the same pattern: navigate internally before going external. This is a pattern learned across 6+ previous WebWalkerQA episodes involving institutional websites, now successfully transferred to a gaming wiki.
\end{tcolorbox}
\end{tcolorbox}

\onecolumn
\section{Algorithm}\label{app:algorithm}
Algorithm~\ref{alg:main} summarizes the full \ourmethod\ pipeline per task, and Algorithm~\ref{alg:evolve} describes the periodic memory evolution.

\begin{algorithm}[h]
\caption{\ourmethod: Per-Task Pipeline}
\label{alg:main}
\begin{algorithmic}[1]
\Require Task $d_q$; memory $\mathcal{G}=(\mathcal{V},\mathcal{E})$; budgets $k_u, k_e, k_s$; dedup threshold $\delta_{\mathrm{dedup}}$
\Ensure Updated memory $\mathcal{G}$
\Statex \textcolor{gray}{\textit{\% Dual-path hyperedge retrieval (\S\ref{sec:episodic_retrieval})}}
\State $\mathcal{P}_0 \gets \textsc{LLM-Decompose}(d_q)$ \Comment{Subtask decomposition}
\State $R_u \gets \operatorname{top\text{-}}k_u$ subtask nodes by $\operatorname{sim}(\boldsymbol{\phi}(\mathcal{P}_0),\, \boldsymbol{\phi}(c_v))$ \hfill $\triangleright$ Eq.~\ref{eq:subtask_retrieval}
\State $\mathcal{E}_{\mathrm{sub}} \gets \{e \in \mathcal{E} \mid V_e \cap R_u \neq \varnothing\}$ \Comment{Subtask path}
\State $\mathcal{E}_{\mathrm{traj}} \gets \operatorname{top\text{-}}k_e$ hyperedges by $\operatorname{sim}(\boldsymbol{\phi}(d_q),\, \mathbf{h}_e)$ \Comment{Trajectory path} \hfill $\triangleright$ Eq.~\ref{eq:task_retrieval}
\State $\mathcal{E}^* \gets \mathcal{E}_{\mathrm{sub}} \cup \mathcal{E}_{\mathrm{traj}}$ \Comment{Fuse} \hfill $\triangleright$ Eq.~\ref{eq:fusion}
\Statex \textcolor{gray}{\textit{\% Skill ranking \& execution (\S\ref{sec:skill_retrieval})}}
\State $R_s \gets \bigcup_{e \in \mathcal{E}^*} V^s_e$; \quad $\kappa(v) \gets |\{e \in \mathcal{E}^* \mid v \in V_e\}|$ for each $v \in R_s$ \hfill $\triangleright$ Eq.~\ref{eq:cooc}
\State $\mathcal{S}_q \gets \operatorname{top\text{-}}k_s$ skills by $\kappa(v)$; \quad $\mathcal{L}_q \gets \{\ell_e \mid e \in \mathcal{E}^*\}$ \hfill $\triangleright$ Eq.~\ref{eq:skill_retrieve}
\State Execute task with context $(\mathcal{S}_q, \mathcal{L}_q)$; observe outcome $r_i$, steps $T_i$ \hfill $\triangleright$ Eq.~\ref{eq:act}
\Statex \textcolor{gray}{\textit{\% Memory update}}
\State Extract skills $S^{\mathrm{new}}_i$ and lesson $\ell_{e_i}$ from trajectory $\tau_i$
\State Deduplicate: merge into existing skill if $\operatorname{sim} \geq \delta_{\mathrm{dedup}}$, else add new node
\State Add hyperedge $e_{\mathrm{new}} = (\mathcal{P}_0 \cup S^{\mathrm{new}}_i,\; d_q,\; \ell_{e_i})$ to $\mathcal{E}$
\State Update utility $\gamma(\cdot)$ via Eq.~\ref{eq:gamma} for all retrieved elements
\end{algorithmic}
\end{algorithm}

\begin{algorithm}[h]
\caption{\ourmethod: Periodic Memory Evolution (every $\lceil 0.1 \times N \rceil$ tasks)}
\label{alg:evolve}
\begin{algorithmic}[1]
\Require Memory $\mathcal{G}=(\mathcal{V},\mathcal{E})$; thresholds $\tau_{\mathrm{prune}}, N_{\min}, \delta_{\mathrm{merge}}$; propagation depth $L$
\Statex \textcolor{gray}{\textit{\% Quality-driven pruning (\S\ref{sec:maintenance})}}
\State Remove all $v$ with $\nu(v) \geq N_{\min}$ and $\gamma(v) < \tau_{\mathrm{prune}}$ from $\mathcal{V}$ and incident hyperedges \hfill $\triangleright$ Eq.~\ref{eq:prune}
\Statex \textcolor{gray}{\textit{\% Structure-informed merging (skill nodes $\mathcal{V}_s$ only)}}
\State $W_{ij} \gets \min(\gamma(v_i),\, \gamma(v_j)) \cdot \sum_{\substack{e \in \mathcal{E} \\ v_i \in V_e,\, v_j \in V_e}} |V_e|^{-1}$ \Comment{Quality-weighted co-occurrence} \hfill $\triangleright$ Eq.~\ref{eq:W}
\State $\tilde{\mathbf{Z}} \gets \mathbf{S}\,\mathbf{Z}$ via $L$-step propagation \Comment{Structural embeddings} \hfill $\triangleright$ Eq.~\ref{eq:propagation}
\For{each pair $(v_a, v_b)$ with $\operatorname{sim}(\tilde{\mathbf{z}}_a,\, \tilde{\mathbf{z}}_b) \geq \delta_{\mathrm{merge}}$}
    \State $v_{\mathrm{new}} \gets \textsc{LLM-Merge}(v_a, v_b)$; reassign hyperedge memberships \hfill $\triangleright$ Eq.~\ref{eq:merge_cand}
\EndFor
\end{algorithmic}
\end{algorithm}

\onecolumn
\section{Prompts}
\label{app:prompts}

\begin{tcolorbox}[
    enhanced,
    colframe=black!60,
    colback=gray!4,
    boxrule=0.8pt,
    arc=3mm,
    title={\textbf{Task Decomposition Prompt (GAIA)}},
    fonttitle=\small\bfseries,
    attach boxed title to top left={yshift=-2mm, xshift=4mm},
    boxed title style={colback=black!60, colframe=black!60, arc=2mm},
    left=4pt, right=4pt, top=4pt, bottom=4pt
]
\small
Decompose this task into 1--3 SHORT steps. Minimize steps to save context window.

\textbf{Available tools:}
\begin{itemize}[nosep,leftmargin=*]
    \item \texttt{web\_search(query)}: Search with SPECIFIC keywords (names, dates, exact phrases)
    \item \texttt{crawl\_page(url)}: Read a URL --- use ONLY when you need specific page content
    \item \texttt{inspect\_file\_as\_text(path)}: Read attached files --- START HERE if task has a file
    \item \texttt{inspect\_file\_as\_image(path)}: View images/figures
    \item \texttt{python code}: Calculations, counting, data processing
\end{itemize}

\textbf{Key principles:}
\begin{itemize}[nosep,leftmargin=*]
    \item If a file is attached, read it FIRST
    \item Use \texttt{web\_search} with the MOST specific terms possible
    \item Prefer python code for calculations over searching
\end{itemize}

\textbf{Task:} \texttt{\{query\}}

Return ONLY a JSON array: \texttt{[\{"name": "action\_name", "description": "use <tool> to <specific action>"\}]}
\end{tcolorbox}

\begin{tcolorbox}[
    enhanced,
    colframe=black!60,
    colback=gray!4,
    boxrule=0.8pt,
    arc=3mm,
    title={\textbf{Task Decomposition Prompt (WebWalkerQA)}},
    fonttitle=\small\bfseries,
    attach boxed title to top left={yshift=-2mm, xshift=4mm},
    boxed title style={colback=black!60, colframe=black!60, arc=2mm},
    left=4pt, right=4pt, top=4pt, bottom=4pt
]
\small
Decompose this web navigation task into 2--4 steps.

\textbf{Available tools:}
\begin{itemize}[nosep,leftmargin=*]
    \item \texttt{web\_search(query)}: Search for specific pages on the website
    \item \texttt{crawl\_page(url)}: Read content from a URL
\end{itemize}

\textbf{Navigation strategy:} ALWAYS start by crawling the root URL to understand site structure. Then navigate to the relevant section (News, Awards, Faculty, etc.) before searching for specifics.

\textbf{Example} for ``Find what award Prof X won in 2021 on university.edu'':

\footnotesize
\begin{verbatim}
[{"name": "explore_site",
  "description": "crawl_page(root_url)
   to find navigation menu"},
 {"name": "navigate_section",
  "description":
   "crawl_page(root_url/awards)"},
 {"name": "find_target",
  "description":
   "web_search('Prof X award 2021
   site:university.edu')"}]
\end{verbatim}
\small

\textbf{Task:} \texttt{\{query\}}

Return ONLY a JSON array.
\end{tcolorbox}

\begin{tcolorbox}[
    enhanced,
    colframe=black!60,
    colback=gray!4,
    boxrule=0.8pt,
    arc=3mm,
    title={\textbf{Task Decomposition Prompt (xBench)}},
    fonttitle=\small\bfseries,
    attach boxed title to top left={yshift=-2mm, xshift=4mm},
    boxed title style={colback=black!60, colframe=black!60, arc=2mm},
    left=4pt, right=4pt, top=4pt, bottom=4pt
]
\small
Decompose this data task into 2--4 steps. Each step must specify WHICH DATA to get and HOW to process it.

\textbf{Available tools:}
\begin{itemize}[nosep,leftmargin=*]
    \item \texttt{web\_search(query)}: Search for data sources
    \item \texttt{crawl\_page(url)}: Read data from specific URLs
    \item \texttt{python code}: Run calculations, unit conversions, comparisons
\end{itemize}

\textbf{Task:} \texttt{\{query\}}

Return ONLY a JSON array: \texttt{[\{"name": "action\_name", "description": "use <tool> to <specific data action>"\}]}
\end{tcolorbox}

\begin{tcolorbox}[
    enhanced,
    colframe=black!60,
    colback=gray!4,
    boxrule=0.8pt,
    arc=3mm,
    title={\textbf{Skill Extraction Prompt (Success)}},
    fonttitle=\small\bfseries,
    attach boxed title to top left={yshift=-2mm, xshift=4mm},
    boxed title style={colback=black!60, colframe=black!60, arc=2mm},
    left=4pt, right=4pt, top=4pt, bottom=4pt
]
\small
Extract reusable knowledge from this SUCCESSFUL task.

\textbf{CRITICAL RULES:}
\begin{itemize}[nosep,leftmargin=*]
    \item Names must be SHORT reusable pattern names (3--5 words), NOT task-specific. \\
          Bad: ``Search for Zip Codes''. Good: ``Targeted Database Search''.
    \item Content must be a COMPLETE paragraph (2--4 sentences) that covers: WHAT to do, WHEN to apply it, and WHY it works.
    \item Extract 1--3 items max. Quality over quantity.
    \item Do NOT include task-specific details (exact names, IDs, URLs). Make knowledge transferable.
\end{itemize}

\textbf{Task:} \texttt{\{query\}} \\
\textbf{Trajectory:} \texttt{\{trajectory\}} \\
\textbf{Result:} \texttt{\{result\}}

Extract skills (what strategies worked well).

Extract as JSON:

\footnotesize
\begin{verbatim}
{
  "skills": [
    {"name": "Short Skill Name",
     "content":
      "2-4 sentence paragraph..."}],
  "knowledge_fragment":
   "ONE OR TWO sentences stating the
    core transferable lesson."
}
\end{verbatim}
\small
\end{tcolorbox}

\begin{tcolorbox}[
    enhanced,
    colframe=black!60,
    colback=gray!4,
    boxrule=0.8pt,
    arc=3mm,
    title={\textbf{Mistake Extraction Prompt (Failure)}},
    fonttitle=\small\bfseries,
    attach boxed title to top left={yshift=-2mm, xshift=4mm},
    boxed title style={colback=black!60, colframe=black!60, arc=2mm},
    left=4pt, right=4pt, top=4pt, bottom=4pt
]
\small
Extract reusable lessons from this FAILED task.

\textbf{CRITICAL RULES:} (same as success extraction)

\textbf{Task:} \texttt{\{query\}} \\
\textbf{Trajectory:} \texttt{\{trajectory\}} \\
\textbf{Result:} \texttt{\{result\}}

Extract mistakes (what went wrong and how to avoid it).

Extract as JSON:
\begin{verbatim}
{
  "mistakes": [{"name": "Short Mistake Name",
                "content": "2-4 sentence paragraph
                 describing what fails, why, and
                 what to do instead."}],
  "knowledge_fragment": "ONE OR TWO sentences
   stating the core failure mode to avoid."
}
\end{verbatim}
\end{tcolorbox}

\begin{tcolorbox}[
    enhanced,
    colframe=black!60,
    colback=gray!4,
    boxrule=0.8pt,
    arc=3mm,
    title={\textbf{Structure-Informed Merging Prompt}},
    fonttitle=\small\bfseries,
    attach boxed title to top left={yshift=-2mm, xshift=4mm},
    boxed title style={colback=black!60, colframe=black!60, arc=2mm},
    left=4pt, right=4pt, top=4pt, bottom=4pt
]
\small
These two \texttt{\{type\}}s are structurally and semantically related. Consolidate them into ONE higher-order \texttt{\{type\}} that captures the knowledge from both.

\texttt{\{TYPE\}} A: \\
\textbf{Name:} \texttt{\{name\_a\}} \\
\textbf{Content:} \texttt{\{content\_a\}}

\texttt{\{TYPE\}} B: \\
\textbf{Name:} \texttt{\{name\_b\}} \\
\textbf{Content:} \texttt{\{content\_b\}}

Create a consolidated \texttt{\{type\}} that:
\begin{itemize}[nosep,leftmargin=*]
    \item Has a SHORT reusable name (3--5 words)
    \item Has content (2--4 sentences) covering BOTH strategies/patterns
    \item Is more general and widely applicable than either individual \texttt{\{type\}}
\end{itemize}

Reply with ONLY JSON: \texttt{\{"name": "...", "content": "..."\}}
\end{tcolorbox}

\begin{tcolorbox}[
    enhanced,
    colframe=black!60,
    colback=gray!4,
    boxrule=0.8pt,
    arc=3mm,
    title={\textbf{Retrieved Memory Format (Agent Context)}},
    fonttitle=\small\bfseries,
    attach boxed title to top left={yshift=-2mm, xshift=4mm},
    boxed title style={colback=black!60, colframe=black!60, arc=2mm},
    left=4pt, right=4pt, top=4pt, bottom=4pt
]
\small
The following memory is assembled once per episode and provided to the agent at the beginning of execution:

\texttt{\#\# Past Experience} \\
\textit{(Distilled lessons from top-$k$ retrieved hyperedges)}
\begin{enumerate}[nosep,leftmargin=*]
    \item[\small 1.] [success] When official exchange data is rate-limited, cross-verified third-party aggregators provide a reliable fallback...
    \item[\small 2.] [failure] Relying on generic search terms instead of querying domain-specific databases yields incomplete data...
\end{enumerate}

\texttt{\#\# Relevant Skills}
\begin{enumerate}[nosep,leftmargin=*]
    \item[\small 1.] \textbf{Targeted Database Search}: Search for specific financial metrics using precise query terms that include the asset, time frame, and required data points...
\end{enumerate}

\texttt{\#\# Mistakes to Avoid}
\begin{enumerate}[nosep,leftmargin=*]
    \item[\small 1.] \textbf{Overreliance on Unverified Aggregators}: Using third-party summaries or social media posts as primary sources for quantitative claims leads to incorrect results...
\end{enumerate}
\end{tcolorbox}

\end{document}